\documentclass{article} % For LaTeX2e
\usepackage[preprint]{colm2026_conference}
\usepackage{microtype}
\usepackage{url}
\usepackage{booktabs, multirow}
\usepackage{amsmath}
\usepackage{amssymb}
\usepackage{amsthm}
\usepackage{lineno}
\usepackage[table]{xcolor}
\usepackage{tcolorbox}
\tcbuselibrary{breakable, skins}
\usepackage{algorithm}
\usepackage{algpseudocode}
\usepackage{mdframed}
\usepackage{wrapfig}
\usepackage{pifont}
\usepackage{caption}
\usepackage{subcaption}
\usepackage{capt-of}
\usepackage{setspace}

\definecolor{phaseblue}{RGB}{30,100,180}
\definecolor{cmtgray}{RGB}{120,130,145}
\definecolor{keycol}{RGB}{180,60,60}
\definecolor{darkblue}{rgb}{0, 0, 0.5}

\usepackage{hyperref}
\hypersetup{
  colorlinks=true,
  citecolor=darkblue,
  linkcolor=darkblue,
  urlcolor=darkblue,
  hypertexnames=false
}

\newcommand{\n}{\textsc{MomentKV}}

\definecolor{momentblue}{RGB}{232, 243, 255}

\title{\n~: Closing the Directional Gap in KV Cache Eviction for Long-Context Inference}

\author{
  Yu Li\textsuperscript{1}, Binxu Li\textsuperscript{2}, Tian Lan\textsuperscript{1}\\
  \textsuperscript{1}George Washington University \\
  \textsuperscript{2}Princeton University \\
  \texttt{\{yul, tlan\}@gwu.edu}
}

\begin{document}
\ifcolmsubmission
\linenumbers
\fi
\maketitle
\begin{abstract}
Autoregressive decoding in Transformer-based language models relies on the KV cache, whose memory footprint grows linearly with sequence length and becomes the primary bottleneck for long-context inference.
KV cache eviction addresses this by retaining a fixed-size subset of key-value pairs and discarding the rest.
We identify that a primary source of output degradation is not the residual attention mass on evicted tokens, which existing methods already minimize, but a directional mismatch between the retained and evicted token sets. 
Specifically, the evicted tokens in practice are often near-orthogonal to the retained ones. Thus, even a small evicted mass could have an oversized impact on the resulting direction distribution and amplify into substantial output error. This reveals a fundamental limit in existing strategies. 
To address this, we propose \n, which maintains compact, small-size moment statistics over the evicted token set, including a count, key mean, value mean, and value-key covariance.
During eviction, the moment statistics is leveraged to identify tokens already well aligned with and captured by the accumulated summary, keeping the evicted set geometrically regular.
During inference, they yield a closed-form first-order approximation of the evicted attention output, forming a mutually reinforcing loop between selective eviction and accurate correction.
On LongBench and RULER with LLaMA-3.1-8B-Instruct and Qwen3-4B-Instruct, \n~outperforms all baselines at every cache budget, with the largest gains under aggressive compression.
\end{abstract}
\vspace{-2mm}
\section{Introduction}
\label{sec:introduction}

Transformer-based language models generate text autoregressively, relying on the KV cache to store previously computed key and value projections and reuse them across decoding steps~\citep{pope2023efficiently}.
While essential for efficient inference, the cache's memory footprint grows with sequence length, quickly dominating GPU memory consumption and I/O bandwidth as context windows scale to tens or hundreds of thousands of tokens~\citep{maurya2024breaking,pope2023efficiently}.
Even with grouped-query attention and reduced precision, the cache for long sequences can reach tens of gigabytes, making KV cache compression a central challenge for deploying long-context language models~\citep{li2024survey,Wang2024ModelTY,li2026right}.

To mitigate this, KV cache eviction is the most widely adopted, retaining a fixed-size subset of KV pairs and discarding the rest.
Existing methods differ primarily in how token importance is estimated for the purpose of eviction, ranging from cumulative attention weights~\citep{Zhang2023H2OHO,liu2023scissorhands} and prefill-phase observation windows~\citep{Li2024SnapKVLK} to layer-wise budget allocation~\citep{Cai2024PyramidKVDK} and value-aware scoring criteria~\citep{devoto2024simple,geng2025accurate,feng2025identify}, demonstrating that retaining a small token subset is sufficient to capture most of the attention mass~\citep{nawrot2024dynamic,zhang2026does,goel2025caote}.
Nevertheless, all eviction methods share the same post-eviction inference procedure: attention is renormalized exclusively over the retained KV pairs, and the evicted ones leave no trace in subsequent operations.

\begin{figure*}[!t]
\centering
\newlength{\figheight}
\setlength{\figheight}{4.3cm}
\begin{minipage}[t]{0.48\textwidth}
  \vspace{0pt}
  \centering
  \includegraphics[height=\figheight]{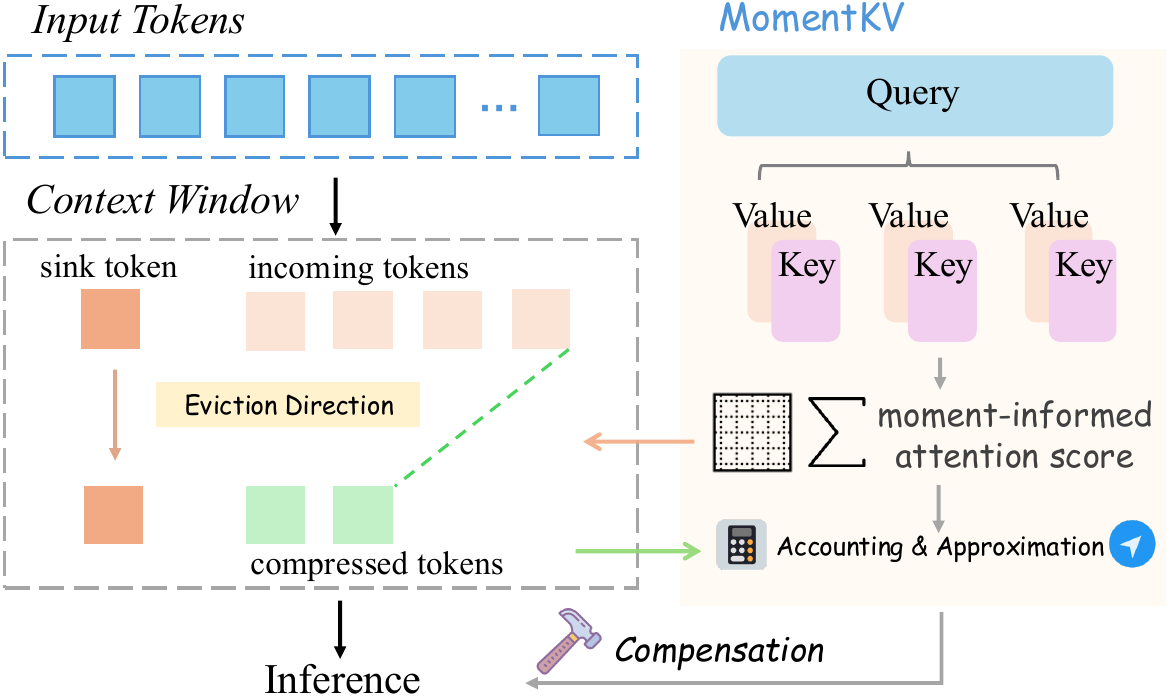}
  \vspace{-4pt}
\subcaption{Overview of \n. Evicted tokens incrementally update compact moment statistics (count, key mean, value mean, and value-key covariance) that guide eviction toward tokens already well captured by the summary and approximate the evicted attention output for post-eviction correction.}
  \label{fig:overview_a}
\end{minipage}%
\hfill
\begin{minipage}[t]{0.45\textwidth}
  \vspace{0pt}
  \centering
  \includegraphics[height=\figheight, width=\textwidth, keepaspectratio]{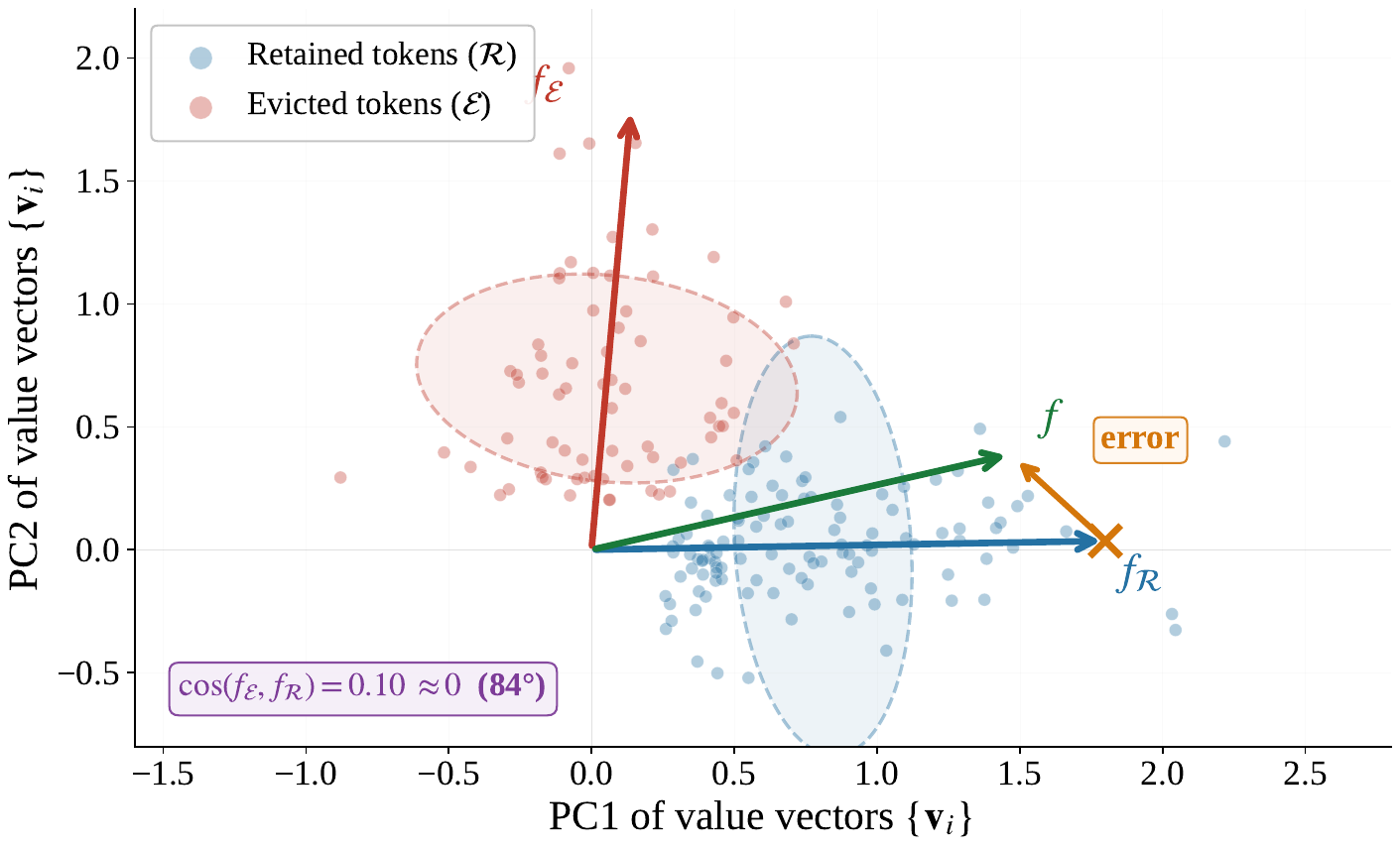}
  \vspace{-4pt}
    \subcaption{PCA of value vectors in one attention head (LLaMA-3-8B, H2O, $N{=}4096$, $L{=}128$). The retained and evicted sub-outputs are near-orthogonal ($\cos\theta \approx 0.10$, $84°$), causing even small evicted attention mass to produce substantial directional error that standard renormalization cannot recover.}
  \label{fig:overview_b}
\end{minipage}
\vspace{-4pt}
\caption{\n~ maintains moment statistics over evicted tokens
to jointly improve eviction decisions
and correct the post-eviction attention output.}
\label{fig:overview}
\vspace{-8pt}
\end{figure*}

When evicted KV pairs carry non-negligible attention mass, this renormalization shifts the attention output toward the subspace spanned by retained value vectors, introducing a systematic directional bias that better scoring cannot resolve~\citep{choromanski2020rethinking, ruan2024towards}.
As shown in Figure~\ref{fig:overview_b}, the retained and evicted sub-outputs are near-orthogonal in practice, so even a small evicted mass produces an oversized impact with substantial relative output error.
The near-orthogonality arises because top-$k$ selection concentrates the retained set on tokens whose keys align with recent query directions, while the evicted set spans a broader, complementary subspace.
Moreover, this creates a dilemma in existing solutions:
a more selective retained set reduces the evicted mass but simultaneously widens potential directional gap due to higher level of concentration, causing existing methods to face diminishing returns as selection quality improves~\citep{hooper2024kvquant,liu2024kivi}.
The path to further improvement therefore lies not only in better token selection but in recovering the directional information that current solutions obliviate~\citep{chang2025palu}.

We propose \n, illustrated in Figure~\ref{fig:overview_a}, which exploits the mass-direction trade-off by maintaining compact summary statistics over the evicted KV pairs.
The statistics are updated upon each eviction and yield a closed-form first-order approximation of the evicted attention output at query time, with small total storage per head independent of context length.
During eviction, the statistics are leveraged to identify tokens whose values are already well captured by the accumulated summary, keeping the evicted set geometrically regular.
During inference, we correct the renormalized output with the moment approximation, restoring an estimate of the lost evicted contribution.
Experiments on LongBench and RULER with LLaMA-3.1-8B-Instruct and Qwen3-4B-Instruct show that \n~consistently outperforms all baselines across cache budgets, with the largest gains under aggressive compression.

\section{Preliminaries}
\label{sec:preliminary}

\textbf{Background.}
The KV cache operates across two phases of inference.
During prefill, the model processes the entire prompt in parallel and populates the cache with key-value pairs for all input tokens~\citep{jin2024compute}.
During decoding, each newly generated token appends one key-value pair to the cache and attends over all cached entries to produce the next token~\citep{dao2023flashattention}.
As decoding progresses, the cache grows by one entry per step per layer, and the attention cost per step increases accordingly~\citep{yuan2024kv,gao2024cost}.

KV cache eviction reduces the cache to a fixed budget of $L$ entries, either once after prefill or continuously during decoding by removing the lowest-scoring entry whenever the cache exceeds $L$.
In either mode, a scoring function assigns an importance value to each cached token, and the lowest-scoring tokens are permanently removed.
Existing scoring functions can be broadly grouped into two families.
Sliding window approaches~\citep{xiao2023efficient} retain a set of initial ``sink'' tokens and the most recent tokens within a fixed window, discarding all intermediate entries without evaluating their content.
Top-$k$ selection methods instead rank all cached tokens by an importance score derived from cumulative attention weights~\citep{Zhang2023H2OHO,liu2023scissorhands}, prefill-phase observation windows~\citep{Li2024SnapKVLK,feng2024adakv}, adaptive budget allocation across layers or heads~\citep{Cai2024PyramidKVDK}, or value-aware criteria that incorporate the geometric properties of value vectors~\citep{devoto2024simple,geng2025accurate,feng2025identify,park2026keydiff}.
Despite differences in scoring, all methods apply the same post-eviction inference rule: attention is renormalized exclusively over the retained set, and the evicted entries are permanently erased.

\textbf{Notation and Formulation.}
To formalize the effect of eviction on the attention output, consider a single attention head with head dimension~$d$.
At a given decoding step, the cache contains $N$ key-value pairs.
Given the current query $\mathbf{q}\in\mathbb{R}^{d}$,
the cached keys $\{\mathbf{k}_{i}\}_{i=1}^{N}$,
and the cached values $\{\mathbf{v}_{i}\}_{i=1}^{N}$,
the attention logits are $s_i = \mathbf{q}^{\top}\mathbf{k}_{i}/\sqrt{d}$
with partition function $Z = \sum_{i=1}^{N}\exp(s_i)$.
The attention weights $\alpha_i = \exp(s_i)/Z$ yield the full attention output, which is the centroid of the cached value vectors under the attention distribution:
\begin{equation}\label{eq:full-attention}
  f(\mathbf{q})
  = \sum_{i=1}^{N} \alpha_i \,\mathbf{v}_{i}
  = \frac{
      \sum_{i=1}^{N}\exp\!\bigl(s_{i}\bigr)\,\mathbf{v}_{i}
    }{Z}.
\end{equation}

After eviction, let $\mathcal{R}\subset[N]$ denote the retained set of size $L$
and $\mathcal{E}=[N]\setminus\mathcal{R}$ the evicted set.
Partitioning the sum over $\mathcal{R}$ and $\mathcal{E}$ and normalizing each part by its own partition function
$Z_{\mathcal{R}}=\sum_{i\in\mathcal{R}}\exp(s_i)$
and $Z_{\mathcal{E}}=\sum_{i\in\mathcal{E}}\exp(s_i)$
yields two sub-outputs:
\begin{equation}\label{eq:sub-outputs}
  f_{\mathcal{R}}(\mathbf{q})
  = \frac{
      \sum_{i\in\mathcal{R}}\exp(s_{i})\,\mathbf{v}_{i}
    }{Z_{\mathcal{R}}},
  \qquad
  f_{\mathcal{E}}(\mathbf{q})
  = \frac{
      \sum_{i\in\mathcal{E}}\exp(s_{i})\,\mathbf{v}_{i}
    }{Z_{\mathcal{E}}}.
\end{equation}
Each sub-output is the attention-weighted centroid of the value vectors restricted to its own token subset.
When the retained and evicted tokens carry value vectors in different regions of $\mathbb{R}^d$, these two centroids can point in substantially different directions, a property that becomes central to the error analysis in Section~\ref{sec:motivation}.

Since $Z = Z_{\mathcal{R}} + Z_{\mathcal{E}}$, the full output decomposes exactly as a convex combination:
\begin{equation}\label{eq:decomposition}
  f(\mathbf{q})
  = w_{\mathcal{R}}\,f_{\mathcal{R}}(\mathbf{q})
  + w_{\mathcal{E}}\,f_{\mathcal{E}}(\mathbf{q}),
  \qquad
  w_{\mathcal{R}} = \frac{Z_{\mathcal{R}}}{Z},\quad
  w_{\mathcal{E}} = \frac{Z_{\mathcal{E}}}{Z},
\end{equation}
where $w_{\mathcal{R}} + w_{\mathcal{E}} = 1$.
The mixing weights correspond to the fraction of the total attention mass assigned to each subset.
Standard eviction outputs $f_{\mathcal{R}}(\mathbf{q})$ alone, which is equivalent to setting $w_{\mathcal{E}}=0$ and redistributing the evicted attention mass entirely onto the retained tokens.
This decomposition separates the eviction error into a scalar factor $w_{\mathcal{E}}$ controlling the magnitude of the lost component and a vector-valued factor $f_{\mathcal{E}}(\mathbf{q}) - f_{\mathcal{R}}(\mathbf{q})$ controlling its direction.
% Existing scoring methods focus exclusively on minimizing $w_{\mathcal{E}}$; Section~\ref{sec:motivation} shows that the directional factor is equally critical and demands a fundamentally different treatment.
\section{The Directional Cost of Token Eviction}
\label{sec:motivation}

As established in Eq.~\eqref{eq:decomposition}, the full attention output decomposes as $f(\mathbf{q}) = w_{\mathcal{R}}\,f_{\mathcal{R}}(\mathbf{q}) + w_{\mathcal{E}}\,f_{\mathcal{E}}(\mathbf{q})$.
Standard eviction discards the evicted component and returns $f_{\mathcal{R}}(\mathbf{q})$ alone.
Subtracting from the full output and using $w_{\mathcal{R}} = 1 - w_{\mathcal{E}}$, the eviction error becomes
\begin{equation}\label{eq:eviction_error}
  f(\mathbf{q}) - f_{\mathcal{R}}(\mathbf{q})
  \;=\;
  w_{\mathcal{E}}
  \;\bigl[\,f_{\mathcal{E}}(\mathbf{q}) - f_{\mathcal{R}}(\mathbf{q})\,\bigr],
\end{equation}
factorizing into a scalar mass term $w_{\mathcal{E}} \in [0,1]$ and a vector-valued divergence $f_{\mathcal{E}}(\mathbf{q}) - f_{\mathcal{R}}(\mathbf{q}) \in \mathbb{R}^d$.
The mass term measures how much attention is lost to eviction, while the divergence measures how differently the two subsets represent the context.

To expose the distinct roles of these two factors, let $\theta = \angle(f_{\mathcal{E}}, f_{\mathcal{R}})$ and decompose the divergence into components parallel and perpendicular to $f_{\mathcal{R}}$:
\begin{equation}\label{eq:parallel_perp}
  f_{\mathcal{E}} - f_{\mathcal{R}}
  \;=\;
  \left(\frac{\|f_{\mathcal{E}}\|}{\|f_{\mathcal{R}}\|}\cos\theta - 1\right) f_{\mathcal{R}}
  \;+\;
  \|f_{\mathcal{E}}\|\sin\theta\;\hat{\mathbf{n}},
\end{equation}
where $\hat{\mathbf{n}}$ is the unit vector orthogonal to $f_{\mathcal{R}}$ in the plane spanned by $f_{\mathcal{E}}$ and $f_{\mathcal{R}}$.
The parallel component rescales the output along $f_{\mathcal{R}}$, and renormalization partially compensates for it.
The perpendicular component, however, shifts the output into a direction entirely outside the span of $f_{\mathcal{R}}$.
Since standard eviction operates exclusively on the retained set, it has no mechanism to recover this orthogonal shift, making the perpendicular error an irreducible cost of eviction under the renormalization-only paradigm.

\begin{figure}[!t]
  \centering
  \includegraphics[width=\textwidth]{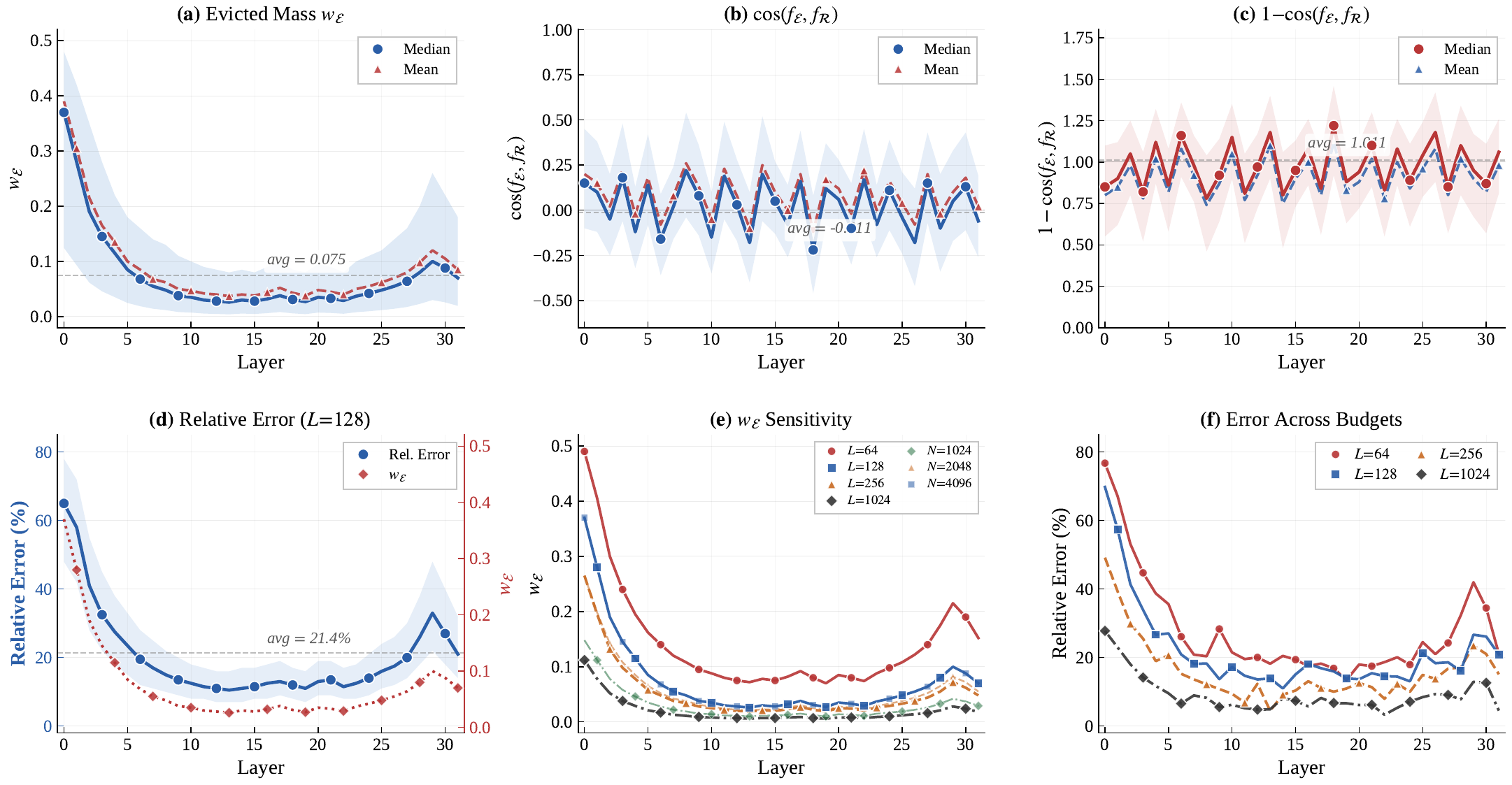}
  \caption{Eviction error analysis on LLaMA-3-8B with H2O selection at $N{=}4096$.
  (a,\,e)~The evicted mass $w_{\mathcal{E}}$ is small,
  averaging below 0.08 at $L{=}128$, and shrinks with larger budgets.
  (b,\,c)~Despite the small mass, the retained and evicted
  sub-outputs remain near-orthogonal across all layers,
  with $\cos(f_{\mathcal{E}}, f_{\mathcal{R}}) \approx 0$.
  (d,\,f)~The near-orthogonality amplifies residual mass
  into over 20\% relative error on average,
  confirming that the directional gap,
  not the evicted mass, is the dominant error source.}
  \label{fig:analysis}
\end{figure}

Substituting into Eq.~\eqref{eq:eviction_error} and using the orthogonality of the two components, the squared error decomposes as
\begin{equation}\label{eq:error_decomp}
  \|f - f_{\mathcal{R}}\|^2
  \;=\;
  w_{\mathcal{E}}^2
  \left[
    \left(\|f_{\mathcal{E}}\|\cos\theta - \|f_{\mathcal{R}}\|\right)^2
    + \|f_{\mathcal{E}}\|^2\sin^2\theta
  \right].
\end{equation}
The parallel error vanishes when $\|f_{\mathcal{E}}\|\cos\theta = \|f_{\mathcal{R}}\|$, but the perpendicular error vanishes only when $\sin\theta = 0$, i.e., when the two sub-outputs are perfectly aligned.
Normalizing by $\|f\|$, the relative error satisfies
\begin{equation}\label{eq:relative_error}
  \frac{\|f - f_{\mathcal{R}}\|}{\|f\|}
  \;=\;
  w_{\mathcal{E}} \cdot \gamma(\theta),
  \qquad
  \gamma(\theta)
  \;=\;
  \frac{\sqrt{(\|f_{\mathcal{E}}\|\cos\theta - \|f_{\mathcal{R}}\|)^2 + \|f_{\mathcal{E}}\|^2\sin^2\theta}}{\|f\|},
\end{equation}
where the amplification coefficient $\gamma(\theta)$ depends on the angle and the sub-output norms, but not on $w_{\mathcal{E}}$.
This multiplicative structure reveals a key insight: even when $w_{\mathcal{E}}$ is small, a large $\gamma(\theta)$ can amplify the residual mass into substantial output error.
We examine both factors empirically on LLaMA-3-8B under H2O selection with $N{=}4096$, as summarized in Figure~\ref{fig:analysis}.

\textbf{Existing selection controls $w_{\mathcal{E}}$ but not $\gamma(\theta)$.}\quad
Figure~\ref{fig:analysis}(a) shows that $w_{\mathcal{E}}$ drops below $0.05$ for most layers at $L{=}128$, with a network-wide average below $0.08$, and Figure~\ref{fig:analysis}(e) confirms that $w_{\mathcal{E}}$ decreases monotonically with larger budgets and shorter contexts.
However, the angle $\theta$ remains persistently large.
Figures~\ref{fig:analysis}(b,c) show that $\cos\theta$ oscillates around zero across all layers, with $1-\cos\theta$ saturated near $1.0$ throughout the network, confirming $\theta \approx \pi/2$.

\textbf{Near-orthogonality produces multiplicative amplification.}\quad
At $\theta = \pi/2$, the parallel term in Eq.~\eqref{eq:error_decomp} reduces to $\|f_{\mathcal{R}}\|^2$ and the perpendicular term equals $\|f_{\mathcal{E}}\|^2$, giving $\gamma(\pi/2) = \sqrt{\|f_{\mathcal{E}}\|^2 + \|f_{\mathcal{R}}\|^2}\,/\,\|f\|$, which remains of order one since both sub-outputs are attention-weighted averages of bounded value vectors.
Figure~\ref{fig:analysis}(d) confirms the amplification empirically: a network-wide average mass below $0.08$ produces over $20\%$ relative error on average, with peaks above $60\%$ in the shallowest layers.
Figure~\ref{fig:analysis}(f) shows the pattern persists across budgets, with shallow layers incurring over $10\%$ error even at $L{=}1024$.

\textbf{Better selection widens the directional gap.}\quad
The near-orthogonality arises because top-$k$ selection concentrates the retained set on tokens whose keys align with recent query directions, while the evicted set accumulates complementary tokens spanning a broader, independent region of the key space~\citep{feng2026defensivekv}.
Improving selection only amplifies this asymmetry: a more selective retained set narrows the directional span of $f_{\mathcal{R}}$ while pushing $f_{\mathcal{E}}$ into an even more complementary subspace, increasing $\gamma(\theta)$ even as $w_{\mathcal{E}}$ decreases.
The two error factors are therefore structurally coupled in opposition, and reducing $w_{\mathcal{E}}$ alone faces diminishing returns.
Addressing this bottleneck requires approximating the perpendicular component of $f_{\mathcal{E}}$ and restoring it during inference.
\section{\n~: Compact Recovery of Evicted Information}
\label{sec:method}

\subsection{First-Order Approximation via Centered Softmax Expansion}
\label{sec:approx}

Recall that the evicted attention output $f_{\mathcal{E}}(\mathbf{q})$ is a softmax-weighted average over the evicted values.
Since the raw logits can be large, we recenter around the mean evicted key $\bar{\mathbf{k}} = n_e^{-1}\sum_{i\in\mathcal{E}}\mathbf{k}_i$ and define centered logits $\delta_{i} = \mathbf{q}^{\top}(\mathbf{k}_{i}-\bar{\mathbf{k}})/\sqrt{d}$, which satisfy $\sum_{i\in\mathcal{E}} \delta_i = 0$ by construction and are typically much smaller in magnitude.
Because softmax is shift-invariant, writing $s_i = \bar{s} + \delta_i$ where $\bar{s} = \mathbf{q}^{\top}\bar{\mathbf{k}}/\sqrt{d}$, the common factor $\exp(\bar{s})$ cancels in the ratio:
\begin{equation}\label{eq:centered_softmax}
  f_{\mathcal{E}}(\mathbf{q})
  \;=\;
  \frac{\sum_{i\in\mathcal{E}} \exp(\delta_i)\,\mathbf{v}_i}
       {\sum_{i\in\mathcal{E}} \exp(\delta_i)},
\end{equation}
leaving $f_{\mathcal{E}}$ expressed entirely in terms of the centered logits.

Applying the first-order Taylor expansion $\exp(\delta_i) \approx 1 + \delta_i$, the zero-sum property $\sum_{i\in\mathcal{E}} \delta_i = 0$ collapses the denominator to $n_e$.
In the numerator, the constant term produces the mean evicted value $\bar{\mathbf{v}} = n_e^{-1}\sum_{i\in\mathcal{E}} \mathbf{v}_i$, and the linear term produces a matrix-vector product involving the empirical value-key covariance $\tilde{\mathbf{S}} = \sum_{i\in\mathcal{E}} \mathbf{v}_i(\mathbf{k}_i - \bar{\mathbf{k}})^{\top} \in \mathbb{R}^{d \times d}$.
Dividing by $n_e$ yields the closed-form approximation
\begin{equation}\label{eq:fhat}
  \hat{f}_{\mathcal{E}}(\mathbf{q})
  \;=\;
  \bar{\mathbf{v}}
  \;+\;
  \frac{1}{n_{e}\sqrt{d}}\,
  \tilde{\mathbf{S}}\,\mathbf{q}.
\end{equation}
The first term $\bar{\mathbf{v}}$ is a query-independent baseline that recovers the mean direction of the evicted sub-output, reducing the angle $\theta$ in Eq.~\eqref{eq:parallel_perp} between the corrected output and the full output.
The second term $\tilde{\mathbf{S}}\,\mathbf{q}$ is a query-adaptive correction that captures the perpendicular variation the baseline alone cannot represent, directly shrinking the $\|f_{\mathcal{E}}\|\sin\theta$ factor in Eq.~\eqref{eq:error_decomp}.
To see why, note that different queries activate different subsets of the evicted keys; the covariance matrix encodes these per-direction correlations between keys and values, allowing the approximation to shift toward the value vectors most relevant to each query.
The approximation error scales as $O(\sigma^2)$ where $\sigma = \max_{i\in\mathcal{E}}|\delta_i|$, and a formal bound is given in Appendix~\ref{app:derivation}.

Crucially, Eq.~\eqref{eq:fhat} depends on the evicted tokens only through four running sums: the count $n_e$, the key sum $\mathbf{s}_k = \sum_{i\in\mathcal{E}}\mathbf{k}_i$, the value sum $\mathbf{s}_v = \sum_{i\in\mathcal{E}}\mathbf{v}_i$, and the outer-product sum $\mathbf{S} = \sum_{i\in\mathcal{E}}\mathbf{v}_i\mathbf{k}_i^{\top}$.
Each statistic is updated by a single addition upon eviction, requiring no storage or revisiting of evicted tokens.
At query time, the means and covariance are recovered via the identity $\tilde{\mathbf{S}} = \mathbf{S} - \mathbf{s}_v\mathbf{s}_k^{\top}/n_e$, and the total storage per head is $O(d^2)$, independent of context length.

\begin{figure}[!t]
\begin{minipage}[t]{0.56\textwidth}
\vspace{0pt}
\begin{algorithm}[H]
\caption{Decoding Step for Attention Head}
\label{alg:momentkv}
\small
\begin{algorithmic}[1]
\Require Retained cache $\mathcal{R}=\{(\mathbf{k}_{i},\mathbf{v}_{i})\}$;
  moment statistics $(n_{e},\,\mathbf{s}_{k},\,\mathbf{s}_{v},\,\mathbf{S})$;
  new token $\mathbf{x}$; budget $L$
\Ensure Attention output $\hat{f}(\mathbf{q})$; updated $\mathcal{R}$ and statistics
\vspace{2pt}
\State $\mathbf{q},\,\mathbf{k}_{\text{new}},\,\mathbf{v}_{\text{new}}
  \leftarrow W_{Q}\mathbf{x},\;W_{K}\mathbf{x},\;W_{V}\mathbf{x}$
\State $\mathcal{R}\leftarrow\mathcal{R}\cup\{(\mathbf{k}_{\text{new}},\,\mathbf{v}_{\text{new}})\}$
\vspace{2pt}
\Statex \textcolor{phaseblue}{\textbf{Phase 1\quad Moment-Informed Eviction}}
\While{$|\mathcal{R}|>L$}
  \State \textbf{if} $n_{e}{>}0$\textbf{:}\;
         $\bar{\mathbf{k}} \leftarrow \mathbf{s}_{k}/n_{e}$,\;
         $\bar{\mathbf{v}} \leftarrow \mathbf{s}_{v}/n_{e}$,\;
         $\tilde{\mathbf{S}} \leftarrow \mathbf{S} - \mathbf{s}_{v}\mathbf{s}_{k}^{\top}/n_{e}$;\;
         \textbf{else}\;
         $\bar{\mathbf{v}},\,\tilde{\mathbf{S}} \leftarrow \mathbf{0}$
  \State Compute $\boldsymbol{\alpha} \leftarrow
    \mathrm{softmax}\!\bigl(\mathbf{q}^{\top}[\mathbf{k}_{j}]_{j\in\mathcal{R}}\,/\,\sqrt{d}\bigr)$
  \For{each $j\in\mathcal{R}$}
    \State $\mathbf{r}_{j} \leftarrow \mathbf{v}_{j} - \bar{\mathbf{v}}
      - \tilde{\mathbf{S}}\,\mathbf{k}_{j}\,/\,(n_{e}\sqrt{d})$
      \hfill\textcolor{cmtgray}{$\triangleright$ moment residual}
    \State $\mathrm{score}(j) \leftarrow \alpha_{j} \cdot \|\mathbf{r}_{j}\|$
      \hfill\textcolor{cmtgray}{$\triangleright$ low $=$ safe to evict}
  \EndFor
  \State Select $j^{*} \leftarrow \arg\min_{j\in\mathcal{R}}\;\mathrm{score}(j)$
  \State Update $n_{e} \leftarrow n_{e} + 1$
  \State Update $\mathbf{s}_{k} \leftarrow \mathbf{s}_{k} + \mathbf{k}_{j^{*}}$,\;\;
         $\mathbf{s}_{v} \leftarrow \mathbf{s}_{v} + \mathbf{v}_{j^{*}}$
  \State Update $\mathbf{S} \leftarrow \mathbf{S} + \mathbf{v}_{j^{*}}\,\mathbf{k}_{j^{*}}^{\top}$
    \hfill\textcolor{cmtgray}{$\triangleright$ rank-one addition}
  \State Remove $\mathcal{R} \leftarrow \mathcal{R}\setminus\{j^{*}\}$
\EndWhile
\vspace{2pt}
\Statex \textcolor{phaseblue}{\textbf{Phase 2\quad Normalization-Corrected Inference}}
\State $f_{\mathcal{R}},\;Z_{\mathcal{R}} \leftarrow \mathrm{Attention}(\mathbf{q},\,\mathcal{R})$
\State \textbf{if} $n_{e}{=}0$ \textbf{then return} $f_{\mathcal{R}}$
  \hfill\textcolor{cmtgray}{$\triangleright$ no evicted tokens yet}
\State $\hat{Z}_{\mathcal{E}} \leftarrow
  n_{e}\cdot\exp\!\bigl(\mathbf{q}^{\top}\bar{\mathbf{k}}\,/\,\sqrt{d}\bigr)$
  \hfill\textcolor{cmtgray}{$\triangleright$ lower bound via Jensen}
\State $\hat{f}_{\mathcal{E}} \leftarrow
  \bar{\mathbf{v}} + \tilde{\mathbf{S}}\,\mathbf{q}\,/\,(n_{e}\sqrt{d})$
  \hfill\textcolor{cmtgray}{$\triangleright$ moment approximation}
\State $\hat{w}_{\mathcal{R}} \leftarrow
  Z_{\mathcal{R}}\,/\,(Z_{\mathcal{R}}+\hat{Z}_{\mathcal{E}})$
  \hfill\textcolor{cmtgray}{$\triangleright$ mixing weight}
\vspace{2pt}
\State \Return $\hat{w}_{\mathcal{R}}\,f_{\mathcal{R}}
  + (1-\hat{w}_{\mathcal{R}})\,\hat{f}_{\mathcal{E}}$
\end{algorithmic}
\end{algorithm}
\end{minipage}%
\hfill
\begin{minipage}[t]{0.42\textwidth}
\vspace{0pt}
\includegraphics[width=\textwidth, height=0.45\textheight, keepaspectratio]{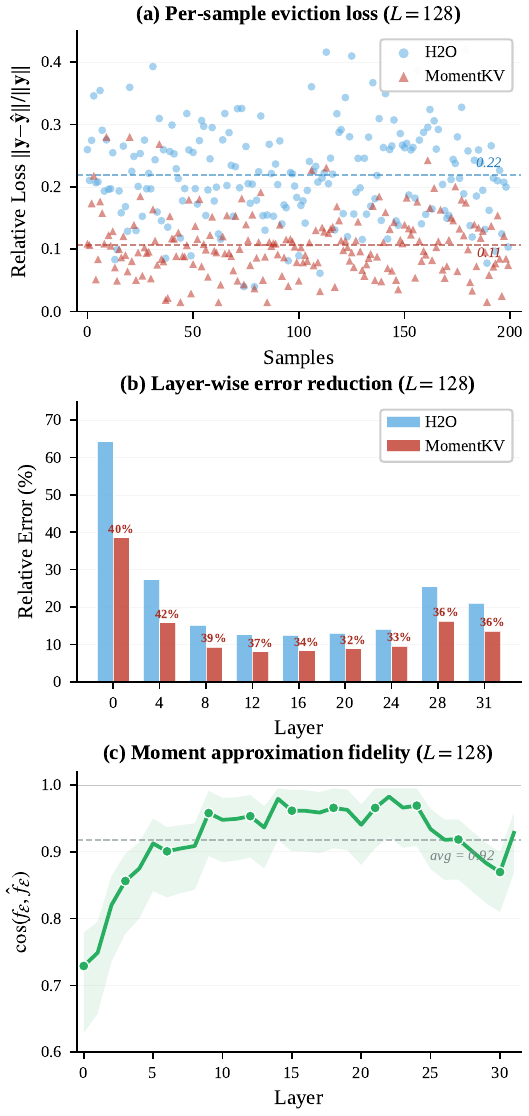}
\vspace{-8pt}
\caption{Empirical validation on LLaMA-3-8B at $L{=}128$ using the Qasper dataset.
(a)~Per-sample eviction loss: H2O vs \n~.
(b)~Layer-wise error and reduction percentage.
(c)~Cosine similarity between $f_{\mathcal{E}}$ and $\hat{f}_{\mathcal{E}}$.}
\label{fig:method_validation}
\end{minipage}
\end{figure}

\subsection{Moment-Informed Eviction and Normalization-Corrected Inference}
\label{sec:dual}

Standard eviction ranks tokens by attention weight alone, ignoring the directional content of their values.
The moment statistics from Eq.~\eqref{eq:fhat} provide a principled way to distinguish tokens whose values are already well captured by the summary from those carrying novel directional information.
For each retained token $j$, we compute a moment residual
\begin{equation}\label{eq:residual}
  \mathbf{r}_j
  \;=\;
  \mathbf{v}_j - \bar{\mathbf{v}}
  - \frac{1}{n_e\sqrt{d}}\,\tilde{\mathbf{S}}\,\mathbf{k}_j,
\end{equation}
measuring the discrepancy between its actual value and the prediction of the affine model evaluated at its own key.
The norm $\|\mathbf{r}_j\|$ quantifies how much novel information token $j$ carries beyond what the accumulated statistics already capture: a small residual indicates that the token's value lies near the affine subspace spanned by $\bar{\mathbf{v}}$ and the column space of $\tilde{\mathbf{S}}$, while a large residual signals directional content that would be lost upon eviction.

The eviction score combines the attention weight with the residual norm as $\mathrm{score}(j) = \alpha_j \cdot \|\mathbf{r}_j\|$, and the token with the lowest score is evicted first.
This multiplicative form ensures that a high-attention token is retained even if its residual is small, because it contributes significantly to $f_{\mathcal{R}}$, while a large-residual token is retained even if its attention weight is low, because evicting it would degrade the moment approximation.
The evicted set therefore gradually accumulates tokens that are well predicted by the moment model, directly suppressing the centered logit spread $\sigma$ and tightening the $O(\sigma^2)$ approximation bound.

Turning to inference, we substitute $\hat{f}_{\mathcal{E}}$ from Eq.~\eqref{eq:fhat} into the decomposition of Eq.~\eqref{eq:decomposition} to recover an estimate of the full attention output.
The retained partition function $Z_{\mathcal{R}}$ is available from the standard attention pass.
For the evicted partition function, we apply Jensen's inequality to the convex exponential function:
\begin{equation}\label{eq:jensen}
  Z_{\mathcal{E}}
  \;=\;
  \sum_{i\in\mathcal{E}} \exp(s_i)
  \;=\;
  n_e \cdot \frac{1}{n_e}\sum_{i\in\mathcal{E}} \exp(\bar{s} + \delta_i)
  \;\geq\;
  n_e \cdot \exp\!\left(\bar{s} + \frac{1}{n_e}\sum_{i\in\mathcal{E}} \delta_i\right)
  \;=\;
  n_e \cdot \exp(\bar{s}),
\end{equation}
where the last equality uses the zero-sum property $\sum_{i\in\mathcal{E}}\delta_i = 0$.
This yields the lower bound $\hat{Z}_{\mathcal{E}} = n_e \cdot \exp(\mathbf{q}^{\top}\bar{\mathbf{k}}/\sqrt{d}) \leq Z_{\mathcal{E}}$, with relative error $O(\sigma^2)$ controlled by the logit variance (see Appendix~\ref{app:jensen}).
With both components estimated, the corrected output takes the form
\begin{equation}\label{eq:fused}
  \hat{f}(\mathbf{q})
  = \hat{w}_{\mathcal{R}}\,f_{\mathcal{R}}(\mathbf{q})
    +
    (1-\hat{w}_{\mathcal{R}})\,\hat{f}_{\mathcal{E}}(\mathbf{q}),
  \qquad
  \hat{w}_{\mathcal{R}}
  = \frac{Z_{\mathcal{R}}}
         {Z_{\mathcal{R}}+\hat{Z}_{\mathcal{E}}}.
\end{equation}
Because $\hat{Z}_{\mathcal{E}}$ underestimates $Z_{\mathcal{E}}$, the weight $\hat{w}_{\mathcal{R}}$ slightly exceeds $w_{\mathcal{R}}$, placing more weight on the exact retained output.
This bias is self-regulating: when $\sigma$ is large, the Jensen gap widens and $\hat{w}_{\mathcal{E}}$ shrinks, limiting the influence of the less accurate approximation; when $\sigma \to 0$, the weights converge to their true values.
Compared to standard eviction, which sets $w_{\mathcal{E}} = 0$ and incurs a relative error of $w_{\mathcal{E}} \cdot \gamma(\theta)$, Eq.~\eqref{eq:fused} reduces both the effective angle $\theta$ and the amplification coefficient $\gamma(\theta)$, shrinking the overall error below what any improvement in $w_{\mathcal{E}}$ alone could achieve.

The two mechanisms form a reinforcing loop: moment-informed eviction keeps $\sigma$ small, tightening both the approximation $\hat{f}_{\mathcal{E}} \approx f_{\mathcal{E}}$ and the Jensen bound on $\hat{Z}_{\mathcal{E}}$, while the improved correction yields more reliable residuals for subsequent eviction decisions.
Figure~\ref{fig:method_validation} provides empirical evidence: \n~ substantially reduces per-sample eviction loss compared to H2O, achieves consistent layer-wise error reduction, and maintains high cosine similarity between the true and approximated evicted outputs across all layers.
The complete procedure is given in Algorithm~\ref{alg:momentkv}.
\section{Results}
\label{sec:experiments}

\subsection{Experimental Setup}
\label{sec:setup}
We evaluate on two benchmarks.
LongBench~\citep{bai2024longbench} is a bilingual multitask benchmark covering single-doc QA, multi-doc QA, summarization, few-shot learning, synthetic tasks and code completion.
RULER~\citep{hsieh2024ruler} complements LongBench by probing retrieval, variable tracking, and multi-hop reasoning under controlled context lengths.

We experiment with two models: LLaMA-3.1-8B-Instruct~\citep{grattafiori2024llama} with 128K context and Qwen3-4B-Instruct-2507~\citep{yang2025qwen3} with 256K context.
Both employ grouped-query attention (GQA) with 32 query heads and 8 KV heads.
We compare against four representative eviction methods: H2O~\citep{Zhang2023H2OHO}, SnapKV~\citep{Li2024SnapKVLK}, PyramidKV~\citep{Cai2024PyramidKVDK} and Ada-KV~\citep{feng2024adakv}.

Our implementation builds upon Ada-KV~\citep{feng2024adakv} and adopts the prefill protocol of SnapKV~\citep{Li2024SnapKVLK}: attention scores are accumulated over a 32-token observation window at the end of the prompt, the first token is retained as an attention sink, adjacent tokens are merged into chunks of size 4 before scoring, and the cache budget is allocated adaptively across heads within each layer based on per-head attention mass.
The cache budget is set to $L \in \{128, 256, 512, 1024\}$, all moment statistics are initialized to zero at the start of each generation, and all experiments are conducted on two NVIDIA H200 GPUs.

\begin{table*}[!t]
\centering
\caption{LongBench results across four cache budgets on two models.
SQA: single-doc QA, MQA: multi-doc QA, Few: few-shot learning.
Avg is computed over all six categories.
\n~achieves the highest average score at every budget level,
with the largest gains under aggressive compression
($+1.35$ over Ada-KV at $L{=}128$ on LLaMA-3.1-8B).
Best in \textbf{bold}, second best \underline{underlined}.}
\label{tab:longbench}
\setlength{\tabcolsep}{5pt}
\resizebox{0.92\textwidth}{!}{%
\begin{tabular}{cl|cccc|cccc}
\toprule
& & \multicolumn{4}{c|}{LLaMA-3.1-8B-Instruct}
  & \multicolumn{4}{c}{Qwen3-4B-Instruct-2507} \\
\cmidrule(lr){3-6}\cmidrule(lr){7-10}
Budget & Method
& SQA & MQA & Few & Avg\,$\uparrow$
& SQA & MQA & Few & Avg\,$\uparrow$ \\
\midrule
-- & Full Cache
& 44.38 & 46.54 & 69.45 & 49.29
& 40.25 & 42.18 & 65.52 & 45.13 \\
\midrule[0.03em]
& H2O
& 34.44 & 41.83 & 58.42 & 42.21
& 33.82 & 38.15 & 57.20 & 40.86 \\
& SnapKV
& 35.98 & 42.85 & 59.57 & 43.21
& 36.90 & 39.48 & 62.85 & 43.24 \\
\multirow{-3}{*}{128}
& PyramidKV
& \underline{37.42} & 43.98 & 64.50 & 44.46
& 36.45 & 39.22 & 62.15 & 42.90 \\
& Ada-KV
& 37.28 & \underline{44.31} & \underline{65.00} & \underline{45.03}
& \underline{37.52} & \underline{39.85} & \underline{63.48} & \underline{43.64} \\
\rowcolor{momentblue}
& \textbf{\n~}
& \textbf{40.15} & \textbf{45.11} & \textbf{66.73} & \textbf{46.38}
& \textbf{38.65} & \textbf{40.72} & \textbf{64.18} & \textbf{44.59} \\
\midrule[0.03em]
& H2O
& 37.78 & 43.18 & 61.26 & 44.28
& 37.15 & 40.02 & 61.85 & 43.16 \\
& SnapKV
& 39.24 & 44.47 & 63.71 & 45.50
& 39.22 & 41.15 & 63.52 & 44.38 \\
\multirow{-3}{*}{256}
& PyramidKV
& \underline{40.71} & 44.34 & 66.94 & 46.01
& 38.85 & 40.88 & 63.10 & 44.11 \\
& Ada-KV
& 40.23 & \underline{45.41} & \underline{67.58} & \underline{46.79}
& \underline{39.68} & \underline{41.42} & \underline{64.15} & \underline{44.71} \\
\rowcolor{momentblue}
& \textbf{\n~}
& \textbf{41.94} & \textbf{45.90} & \textbf{68.42} & \textbf{47.59}
& \textbf{39.95} & \textbf{41.85} & \textbf{64.82} & \textbf{45.23} \\
\midrule[0.03em]
& H2O
& 40.83 & 44.50 & 63.66 & 46.08
& 39.55 & 41.42 & 63.78 & 44.42 \\
& SnapKV
& 41.77 & 45.36 & 67.60 & 47.29
& 39.85 & 41.72 & 64.38 & 44.86 \\
\multirow{-3}{*}{512}
& PyramidKV
& \underline{42.58} & 45.55 & 68.23 & 47.41
& 39.62 & 41.55 & 64.05 & 44.63 \\
& Ada-KV
& 41.60 & \underline{45.60} & \underline{68.50} & \underline{47.55}
& \underline{40.05} & \underline{41.88} & \underline{64.92} & \underline{45.10} \\
\rowcolor{momentblue}
& \textbf{\n~}
& \textbf{43.41} & \textbf{46.17} & \textbf{69.13} & \textbf{48.27}
& \textbf{40.18} & \textbf{42.05} & \textbf{65.25} & \textbf{45.45} \\
\midrule[0.03em]
& H2O
& 43.40 & 45.84 & 67.81 & 48.00
& 40.02 & 41.95 & 65.12 & 44.94 \\
& SnapKV
& \underline{43.50} & 46.11 & 68.60 & 48.30
& 40.12 & 42.05 & 65.28 & 45.23 \\
\multirow{-3}{*}{1024}
& PyramidKV
& 43.34 & 46.23 & 68.46 & 48.09
& 39.95 & 41.92 & 65.02 & 45.07 \\
& Ada-KV
& 43.42 & \underline{46.32} & \underline{68.84} & \underline{48.30}
& \underline{40.22} & \underline{42.12} & \underline{65.42} & \underline{45.33} \\
\rowcolor{momentblue}
& \textbf{\n~}
& \textbf{44.28} & \textbf{46.46} & \textbf{69.43} & \textbf{48.89}
& \textbf{40.24} & \textbf{42.15} & \textbf{65.48} & \textbf{45.38} \\
\bottomrule
\end{tabular}%
}
\end{table*}

\subsection{Main Results}
\label{sec:main_results}

Table~\ref{tab:longbench} reports the LongBench results.
\n~achieves the highest average score at every budget level on both models.
On LLaMA-3.1-8B, it surpasses the strongest baseline Ada-KV by $+1.35$ at $L{=}128$, retaining 94.1\% of the full-cache performance.
The gains are most pronounced on QA and few-shot tasks, which require attending to specific facts scattered across the full context and are therefore most affected by the permanent information loss that standard eviction introduces.
Summarization and code tasks show smaller margins, as they rely more on local context that is likely retained in the cache.
On Qwen3-4B, \n~leads consistently with the largest margin at $L{=}128$, confirming that the benefit generalizes across model architectures and scales.

As shown in Figure~\ref{fig:budget}, the advantage narrows at larger budgets, with the margin over Ada-KV decreasing to $+0.59$ at $L{=}1024$.
This trend follows directly from Eq.~\eqref{eq:eviction_error}: a larger budget reduces the evicted mass $w_\mathcal{E}$ and leaves less directional error available for correction.
Even at $L{=}1024$ where all baselines converge to within 1 point of each other, \n~still maintains a consistent lead, suggesting that the moment correction captures information that improved selection alone cannot recover.

Table~\ref{tab:ruler} reports RULER results at $L{=}128$.
\n~outperforms Ada-KV by $+3.4$ on LLaMA-3.1-8B and $+3.5$ on Qwen3-4B, proportionally larger margins than on LongBench.
RULER tasks require precise retrieval of specific tokens embedded at controlled positions, making them especially sensitive to the loss of evicted information and validating that the moment correction is most valuable in retrieval-intensive settings.

\vspace{-2mm}
\begin{wraptable}{r}{0.42\textwidth}
\vspace{-14pt}
\centering
\setlength{\tabcolsep}{2.5pt}
\caption{RULER scores (16K, $L{=}128$). \n~leads by $+3.4$ and $+3.5$ over Ada-KV.}
\label{tab:ruler}
\vspace{-8pt}
\begin{tabular}{l|cc}
\toprule
Method & LLaMA-3.1 & Qwen3-4B \\
\midrule
Full Cache   & 85.2 & 80.5 \\
\midrule
H2O          & 62.8 & 58.5 \\
SnapKV       & 71.5 & 66.2 \\
PyramidKV    & 70.2 & 65.5 \\
Ada-KV       & \underline{73.8} & \underline{68.5} \\
\rowcolor{momentblue}
\textbf{\n~}
             & \textbf{77.2} & \textbf{72.0} \\
\bottomrule
\end{tabular}
\vspace{0pt}
\includegraphics[width=\linewidth]{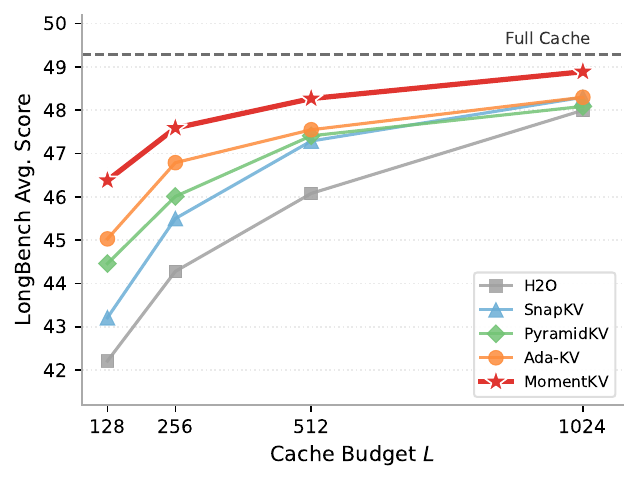}
\vspace{-16pt}
\captionof{figure}{LongBench avg.\ score across cache budgets on LLaMA-3.1-8B. The advantage is largest at small $L$ and narrows as the budget grows.}
\label{fig:budget}
\vspace{-12pt}
\end{wraptable}

\subsection{Ablation and Further Analysis}
\label{sec:ablation}

Table~\ref{tab:ablation_component} isolates the two components of \n, with SnapKV as the baseline.
Moment-informed eviction (MI) alone yields $+0.57$ at $L{=}128$, while normalization-corrected inference (NC) alone contributes $+2.41$, confirming that restoring the evicted output direction is more impactful than refining selection.
The ${\sim}4{\times}$ ratio is consistent with the analysis in Section~\ref{sec:motivation}, where the dominant error source is the amplification coefficient $\gamma(\theta)$ that only NC addresses.
Combining both achieves $+3.17$, exceeding the sum of individual gains by $+0.19$ due to the reinforcing loop described in Section~\ref{sec:dual}.

Table~\ref{tab:ablation_order} compares approximation orders and their overhead.
The zeroth-order variant (mean value $\bar{\mathbf{v}}$ only, $O(d)$ per head) already improves over SnapKV by $+0.94$, showing that even a query-independent correction recovers useful evicted information.
The first-order approximation adds $O(d^2)$ storage per head (4.1\,MB total, 6.4\% of retained cache) and 1.6\,ms latency, yet more than triples the zeroth-order gain by capturing query-dependent directional variation.
Both costs are determined by $d$ and do not grow with context length, making \n~practical for arbitrarily long sequences while reducing per-token latency by over 65\% compared to the full cache.
\begin{table}[!t]
\centering
\begin{minipage}[t]{0.44\linewidth}
\vspace{0pt}
\centering
\parbox{\linewidth}{\caption{Component ablation on LongBench avg.\ score (LLaMA-3.1-8B). MI: moment-informed eviction, NC: normalization-corrected inference.}\label{tab:ablation_component}}
\vspace{1pt}
\setlength{\tabcolsep}{4pt}
\renewcommand{\arraystretch}{1.1}
\resizebox{\textwidth}{!}{
\begin{tabular}{cc|ccc}
\toprule
MI & NC & $L{=}128$ & $L{=}256$ & $L{=}512$ \\
\midrule
            &              & 43.21 & 45.50 & 47.29 \\
\checkmark  &              & 43.78 {\scriptsize\color{teal}+0.57} & 45.85 {\scriptsize\color{teal}+0.35} & 47.52 {\scriptsize\color{teal}+0.23} \\
            & \checkmark   & 45.62 {\scriptsize\color{teal}+2.41} & 47.05 {\scriptsize\color{teal}+1.55} & 47.98 {\scriptsize\color{teal}+0.69} \\
\rowcolor{momentblue}
\checkmark  & \checkmark   & \textbf{46.38} {\scriptsize\color{teal}+3.17} & \textbf{47.59} {\scriptsize\color{teal}+2.09} & \textbf{48.27} {\scriptsize\color{teal}+0.98} \\
\bottomrule
\end{tabular}}
\end{minipage}%
\hfill
\begin{minipage}[t]{0.54\linewidth}
\vspace{0pt}
\centering
\parbox{\linewidth}{\caption{Approximation order and overhead on LongBench avg.\ score (LLaMA-3.1-8B, $L{=}128$). Baseline: SnapKV.}\label{tab:ablation_order}}
\vspace{1pt}
\setlength{\tabcolsep}{3pt}
\renewcommand{\arraystretch}{1.1}
\resizebox{\textwidth}{!}{
\begin{tabular}{l|cccc}
\toprule
Approx. & Avg. & $\Delta$ & Aux.\ Mem. & Lat./tok \\
\midrule
None (SnapKV)
  & 43.21 & {--}   & 0          & 8.2\,ms \\
0th ($\bar{\mathbf{v}}$)
  & 44.15 & +0.94  & $2d$       & 8.4\,ms \\
\rowcolor{momentblue}
\textbf{1st (ours)}
  & \textbf{46.38} & \textbf{+3.17} & $d^2{+}2d$ & 9.8\,ms \\
\midrule
Full Cache
  & 49.29 & {--}   & Full       & 28.5\,ms \\
\bottomrule
\end{tabular}}
\end{minipage}
\end{table}

\section{Conclusion}

We have shown that a key source of quality loss in KV cache eviction is not insufficient token selection but the directional mismatch between retained and evicted sub-outputs, which renormalization-only inference cannot fully address.
\n~closes this gap by maintaining compact moment statistics over the evicted set that serve a dual role: guiding eviction toward tokens already well captured by the accumulated summary, and providing a closed-form first-order correction to the post-eviction attention output.
The two mechanisms form a reinforcing loop, as moment-informed eviction suppresses the centered logit spread $\sigma$ that controls the approximation error, while accurate correction yields more reliable residuals for subsequent eviction decisions.
Experiments on LongBench and RULER with two model families confirm consistent improvements at every cache budget, with the largest gains under aggressive compression in retrieval-intensive tasks.
Future directions include per-head adaptive approximation orders and integration with orthogonal techniques such as quantization and low-rank factorization.
\bibliography{ref}
\bibliographystyle{colm2026_conference}

\newpage
\appendix
\section*{Appendix}

\section{Derivation of the First-Order Approximation}
\label{app:derivation}

We derive Eq.~\eqref{eq:fhat} in full detail, describe the incremental maintenance of the underlying sufficient statistics, and provide a quantitative bound on the approximation error.

\subsection{Centered Softmax and First-Order Expansion}
\label{app:expansion}

The softmax function is invariant to additive shifts: for any constant $c$,
\begin{equation}
    \frac{\exp(s_i + c)}{\sum_j \exp(s_j + c)}
    = \frac{\exp(s_i)}{\sum_j \exp(s_j)}.
\end{equation}
Define the mean evicted key $\bar{\mathbf{k}} = \frac{1}{n_e}\sum_{i \in \mathcal{E}} \mathbf{k}_i$, the mean logit $\bar{s}(\mathbf{q}) = \mathbf{q}^\top \bar{\mathbf{k}} / \sqrt{d}$, and the centered logit
\begin{equation}
    \delta_i(\mathbf{q})
    = \frac{\mathbf{q}^\top (\mathbf{k}_i - \bar{\mathbf{k}})}{\sqrt{d}}
    = s_i(\mathbf{q}) - \bar{s}(\mathbf{q}).
\end{equation}
By construction,
\begin{equation}
    \sum_{i \in \mathcal{E}} \delta_i(\mathbf{q})
    = \frac{\mathbf{q}^\top}{\sqrt{d}}
      \sum_{i \in \mathcal{E}} (\mathbf{k}_i - \bar{\mathbf{k}})
    = 0
    \label{eq:app_zero_sum}
\end{equation}
for every query $\mathbf{q}$.
The zero-sum property is the key structural ingredient that makes the first-order expansion exact in the denominator and confines all approximation error to second-order terms.
Since $s_i = \bar{s} + \delta_i$, the common factor $\exp(\bar{s})$ cancels in the softmax ratio:
\begin{equation}
    f_{\mathcal{E}}(\mathbf{q})
    = \frac{\sum_{i \in \mathcal{E}} \exp(\delta_i)\, \mathbf{v}_i}
           {\sum_{i \in \mathcal{E}} \exp(\delta_i)}.
    \label{eq:app_centered}
\end{equation}

We now apply the Taylor expansion $\exp(\delta_i) = 1 + \delta_i + \frac{1}{2}\delta_i^2 e^{\xi_i}$ for some $\xi_i$ between $0$ and $\delta_i$, retaining only first-order terms.
Let $\sigma = \max_{i \in \mathcal{E}} |\delta_i|$ denote the maximum centered logit magnitude, which controls the quality of the approximation throughout.

For the denominator, the zero-sum property~\eqref{eq:app_zero_sum} eliminates the linear term:
\begin{equation}
    \sum_{i \in \mathcal{E}} \exp(\delta_i)
    = n_e + \underbrace{\sum_{i \in \mathcal{E}} \delta_i}_{=\,0} + R_{\mathrm{den}}
    = n_e + R_{\mathrm{den}},
    \label{eq:app_denom}
\end{equation}
where $|R_{\mathrm{den}}| \leq \frac{1}{2} e^\sigma \sum_{i \in \mathcal{E}} \delta_i^2 \leq \frac{n_e \sigma^2}{2} e^\sigma$.
Without the zero-sum property, the denominator would contain an $O(\sigma)$ linear term, and the overall approximation would degrade to $O(\sigma)$ rather than $O(\sigma^2)$.

For the numerator, expanding and separating terms:
\begin{equation}
    \sum_{i \in \mathcal{E}} \exp(\delta_i)\, \mathbf{v}_i
    = \underbrace{\sum_{i \in \mathcal{E}} \mathbf{v}_i}_{= \, n_e \bar{\mathbf{v}}}
      + \sum_{i \in \mathcal{E}} \delta_i\, \mathbf{v}_i
      + \mathbf{R}_{\mathrm{num}},
    \label{eq:app_numer}
\end{equation}
where $\bar{\mathbf{v}} = \frac{1}{n_e}\sum_{i \in \mathcal{E}} \mathbf{v}_i$ and $\|\mathbf{R}_{\mathrm{num}}\| \leq \frac{n_e \sigma^2 V_{\max}}{2} e^\sigma$ with $V_{\max} = \max_{i \in \mathcal{E}} \|\mathbf{v}_i\|$.
Unlike the denominator, the linear term in the numerator does not vanish because the $\delta_i$ are weighted by distinct value vectors $\mathbf{v}_i$.
It is precisely this surviving linear term that provides the query-adaptive correction in the final approximation.

Substituting $\delta_i = \mathbf{q}^\top(\mathbf{k}_i - \bar{\mathbf{k}})/\sqrt{d}$ into the linear term and factoring:
\begin{align}
    \sum_{i \in \mathcal{E}} \delta_i\, \mathbf{v}_i
    &= \frac{1}{\sqrt{d}}
       \sum_{i \in \mathcal{E}}
       \mathbf{v}_i\, (\mathbf{k}_i - \bar{\mathbf{k}})^\top \mathbf{q}
    = \frac{1}{\sqrt{d}}
       \left(\sum_{i \in \mathcal{E}}
       \mathbf{v}_i (\mathbf{k}_i - \bar{\mathbf{k}})^\top\right) \mathbf{q}
    = \frac{\tilde{\mathbf{S}}}{\sqrt{d}}\, \mathbf{q},
\end{align}
where $\tilde{\mathbf{S}} = \sum_{i \in \mathcal{E}} \mathbf{v}_i(\mathbf{k}_i - \bar{\mathbf{k}})^\top \in \mathbb{R}^{d \times d}$ is the empirical cross-covariance between evicted values and centered keys.
The matrix $\tilde{\mathbf{S}}$ encodes how the value vectors co-vary with key directions across the evicted set: for a query $\mathbf{q}$ aligned with a particular key direction, the product $\tilde{\mathbf{S}}\mathbf{q}$ retrieves the corresponding value component, enabling the approximation to adapt to different query directions.
The numerator approximation thus becomes
\begin{equation}
    \sum_{i \in \mathcal{E}} \exp(\delta_i)\, \mathbf{v}_i
    \approx n_e \bar{\mathbf{v}} + \frac{\tilde{\mathbf{S}}}{\sqrt{d}}\, \mathbf{q}.
    \label{eq:app_numer_result}
\end{equation}
Dividing Eq.~\eqref{eq:app_numer_result} by $n_e$ from Eq.~\eqref{eq:app_denom} yields
\begin{equation}
    \hat{f}_{\mathcal{E}}(\mathbf{q})
    = \bar{\mathbf{v}}
      + \frac{1}{n_e \sqrt{d}}\, \tilde{\mathbf{S}}\, \mathbf{q},
    \label{eq:app_fhat}
\end{equation}
which is Eq.~\eqref{eq:fhat}.
The approximation is an affine function of the query: the intercept $\bar{\mathbf{v}}$ captures the query-independent mean of the evicted sub-output, and the linear term $\tilde{\mathbf{S}}\mathbf{q}/(n_e\sqrt{d})$ captures the first-order query-dependent variation around the mean.

The matrix $\tilde{\mathbf{S}}$ can be computed from four running sums without revisiting any evicted token.
Define the sufficient statistics
\begin{equation}
    n_e = |\mathcal{E}|,
    \quad
    \mathbf{s}_k = \sum_{i \in \mathcal{E}} \mathbf{k}_i,
    \quad
    \mathbf{s}_v = \sum_{i \in \mathcal{E}} \mathbf{v}_i,
    \quad
    \mathbf{S} = \sum_{i \in \mathcal{E}} \mathbf{v}_i \mathbf{k}_i^\top.
    \label{eq:app_sufficient}
\end{equation}
Using $\mathbf{s}_v = n_e \bar{\mathbf{v}}$ and $\bar{\mathbf{k}} = \mathbf{s}_k / n_e$:
\begin{align}
    \tilde{\mathbf{S}}
    &= \sum_{i \in \mathcal{E}} \mathbf{v}_i (\mathbf{k}_i - \bar{\mathbf{k}})^\top
    = \sum_{i \in \mathcal{E}} \mathbf{v}_i \mathbf{k}_i^\top
       - \left(\sum_{i \in \mathcal{E}} \mathbf{v}_i\right) \bar{\mathbf{k}}^\top
    = \mathbf{S} - \frac{\mathbf{s}_v \mathbf{s}_k^\top}{n_e},
    \label{eq:app_S_tilde}
\end{align}
which is analogous to the standard computational formula $\mathrm{Cov}(X,Y) = \mathbb{E}[XY] - \mathbb{E}[X]\mathbb{E}[Y]$, applied to the empirical distribution over evicted tokens.
At query time, the full set of inference quantities is recovered as
\begin{equation}
    \bar{\mathbf{k}} = \frac{\mathbf{s}_k}{n_e},
    \qquad
    \bar{\mathbf{v}} = \frac{\mathbf{s}_v}{n_e},
    \qquad
    \tilde{\mathbf{S}} = \mathbf{S} - \frac{\mathbf{s}_v \mathbf{s}_k^\top}{n_e}.
\end{equation}
When token $j^*$ is evicted from the retained set, the statistics are updated by simple addition:
\begin{equation}
    n_e \mathrel{+}= 1,
    \quad
    \mathbf{s}_k \mathrel{+}= \mathbf{k}_{j^*},
    \quad
    \mathbf{s}_v \mathrel{+}= \mathbf{v}_{j^*},
    \quad
    \mathbf{S} \mathrel{+}= \mathbf{v}_{j^*} \mathbf{k}_{j^*}^\top.
    \label{eq:app_update}
\end{equation}
Each update consists of a rank-one addition to $\mathbf{S}$ and vector additions to $\mathbf{s}_k$ and $\mathbf{s}_v$, at $O(d^2)$ cost per eviction step.
The update is purely additive: no evicted token needs to be stored or revisited after incorporation into the statistics, and the cost is independent of $n_e$.

The scalar $n_e$ is negligible in storage.
The vectors $\mathbf{s}_k, \mathbf{s}_v \in \mathbb{R}^d$ together require $2d$ values, and the dominant cost is the matrix $\mathbf{S} \in \mathbb{R}^{d \times d}$, requiring $d^2$ values.
For $d = 128$ with 16-bit precision, $\mathbf{S}$ occupies $128 \times 128 \times 2 = 32{,}768$ bytes $\approx 32$\,KB per head, equivalent to 64 KV pairs.
Including $\mathbf{s}_k$ and $\mathbf{s}_v$, the total per-head overhead is equivalent to roughly 128 KV pairs, independent of the context length $N$.
For a model with $L_{\text{layers}}$ layers and $H$ KV heads per layer, the total overhead is $L_{\text{layers}} \times H \times (d^2 + 2d) \times 2$ bytes.
Table~\ref{tab:storage_models} reports concrete values for the two experimental models.

\begin{table}[h]
\centering
\caption{Moment statistics storage overhead. Retained cache size computed at $L{=}128$ with 16-bit precision.}
\label{tab:storage_models}
\setlength{\tabcolsep}{5pt}
\renewcommand{\arraystretch}{1.1}
\begin{tabular}{l|ccccc}
\toprule
Model & $L_{\text{layers}}$ & $H$ & $d$ & Moment Overhead & \% of Retained Cache \\
\midrule
LLaMA-3.1-8B  & 32 & 8 & 128 & 8.5\,MB  & 6.4\% \\
Qwen3-4B      & 36 & 8 & 128 & 9.6\,MB  & 7.2\% \\
\bottomrule
\end{tabular}
\end{table}

The overhead is modest in both absolute and relative terms.
As the cache budget $L$ increases, the relative overhead decreases proportionally since the moment storage is fixed at $O(d^2)$ per head while the retained cache grows as $O(Ld)$.
At $L{=}1024$, the moment statistics account for less than 1\% of the total memory footprint.

\subsection{Quantitative Error Bound}
\label{app:error_bound}

We bound the gap $\|f_{\mathcal{E}}(\mathbf{q}) - \hat{f}_{\mathcal{E}}(\mathbf{q})\|$.
Write the exact numerator and denominator of Eq.~\eqref{eq:app_centered} as
\begin{equation}
    A = n_e \bar{\mathbf{v}} + \frac{\tilde{\mathbf{S}}\,\mathbf{q}}{\sqrt{d}} + \mathbf{R}_{\mathrm{num}},
    \qquad
    B = n_e + R_{\mathrm{den}},
\end{equation}
so that $f_{\mathcal{E}} = A / B$.
Expanding $B^{-1} = n_e^{-1}(1 + R_{\mathrm{den}} / n_e)^{-1}$ to first order:
\begin{align}
    \frac{A}{B}
    &= \frac{A}{n_e}\left(1 - \frac{R_{\mathrm{den}}}{n_e} + O\!\left(\frac{R_{\mathrm{den}}^2}{n_e^2}\right)\right) \notag\\
    &= \frac{A}{n_e} - \frac{A \cdot R_{\mathrm{den}}}{n_e^2} + O(\sigma^4).
\end{align}
Since $\hat{f}_{\mathcal{E}} = (A - \mathbf{R}_{\mathrm{num}}) / n_e$, the difference is
\begin{align}
    f_{\mathcal{E}} - \hat{f}_{\mathcal{E}}
    &= \frac{\mathbf{R}_{\mathrm{num}}}{n_e}
      - \frac{A \cdot R_{\mathrm{den}}}{n_e^2}
      + O(\sigma^4).
\end{align}
The first term arises from the numerator remainder (the second-order Taylor residual weighted by value vectors), and the second from the denominator remainder propagated through the division.
Taking norms and substituting the remainder bounds from Eqs.~\eqref{eq:app_denom}--\eqref{eq:app_numer}:
\begin{align}
    \|f_{\mathcal{E}}(\mathbf{q}) - \hat{f}_{\mathcal{E}}(\mathbf{q})\|
    &\leq \frac{\|\mathbf{R}_{\mathrm{num}}\|}{n_e}
      + \frac{\|A\| \cdot |R_{\mathrm{den}}|}{n_e^2}
      + O(\sigma^4) \notag\\
    &\leq \frac{\sigma^2 e^\sigma}{2} V_{\max}
      + \frac{(\|\hat{f}_{\mathcal{E}}\| + \frac{\|\mathbf{R}_{\mathrm{num}}\|}{n_e}) \cdot \frac{n_e \sigma^2}{2} e^\sigma}{n_e^2}
      + O(\sigma^4) \notag\\
    &= \frac{\sigma^2 e^\sigma}{2}
      \left(V_{\max} + \|\hat{f}_{\mathcal{E}}\|\right)
      + O(\sigma^4)
    = O(\sigma^2).
    \label{eq:app_error_bound}
\end{align}

The bound admits a geometric interpretation.
The Cauchy--Schwarz inequality gives $|\delta_i| = |\mathbf{q}^\top(\mathbf{k}_i - \bar{\mathbf{k}})| / \sqrt{d} \leq \|\mathbf{q}\| \cdot \|\mathbf{k}_i - \bar{\mathbf{k}}\| / \sqrt{d}$,
so $\sigma \leq \|\mathbf{q}\| \cdot r_{\max} / \sqrt{d}$ where $r_{\max} = \max_{i \in \mathcal{E}} \|\mathbf{k}_i - \bar{\mathbf{k}}\|$ is the maximum key deviation from the centroid.
Substituting into Eq.~\eqref{eq:app_error_bound}:
\begin{equation}
    \|f_{\mathcal{E}} - \hat{f}_{\mathcal{E}}\|
    \leq \frac{\|\mathbf{q}\|^2}{2d}\,
    e^{\|\mathbf{q}\| r_{\max} / \sqrt{d}}\,
    r_{\max}^2\,
    \left(V_{\max} + \|\hat{f}_{\mathcal{E}}\|\right)
    + O(\sigma^4).
    \label{eq:app_spectral_bound}
\end{equation}
Three factors control approximation quality.
First, the key spread $r_{\max}$: the approximation is tightest when evicted keys cluster around their centroid $\bar{\mathbf{k}}$, and the moment-informed eviction criterion in Section~\ref{sec:dual} directly promotes small $r_{\max}$ by preferentially evicting tokens whose keys are close to $\bar{\mathbf{k}}$ and whose values are well predicted by the affine model.
Second, the query norm $\|\mathbf{q}\|$: larger query norms amplify the centered logits, but in practice query norms are bounded by the layer normalization applied before the attention projection.
Third, the head dimension $d$: the $1/d$ factor reflects the $1/\sqrt{d}$ scaling applied to attention logits, and for the standard $d{=}128$ used in both experimental models, this provides approximately an $11{\times}$ reduction in logit magnitude compared to unnormalized dot products.

The moment residual $\mathbf{r}_j = \mathbf{v}_j - \bar{\mathbf{v}} - \tilde{\mathbf{S}}\mathbf{k}_j/(n_e\sqrt{d})$ used in the eviction score (Eq.~\ref{eq:residual}) connects directly to the error bound.
A token with small $\|\mathbf{r}_j\|$ has its value well predicted by the affine model, meaning that adding it to the evicted set increases $n_e$ without substantially increasing $V_{\max}$ or $r_{\max}$ in Eq.~\eqref{eq:app_spectral_bound}.
Conversely, a token with large $\|\mathbf{r}_j\|$ would introduce a value vector that deviates from the affine subspace, potentially increasing both $\|\mathbf{R}_{\mathrm{num}}\|$ and $r_{\max}$.
The multiplicative score $\alpha_j \cdot \|\mathbf{r}_j\|$ therefore jointly minimizes the impact on both the retained output quality (through $\alpha_j$) and the approximation quality (through $\|\mathbf{r}_j\|$).

\section{Jensen Bound on the Evicted Partition Function}
\label{app:jensen}

We derive the estimate $\hat{Z}_{\mathcal{E}} = n_e \exp(\bar{s})$ used in Eq.~\eqref{eq:fused}, analyze its tightness, and characterize the self-regulating behavior of the resulting mixing weight.

\subsection{Derivation and Tightness}
\label{app:jensen_derivation}

The true evicted partition function is $Z_{\mathcal{E}} = \sum_{i \in \mathcal{E}} \exp(s_i)$.
Writing $s_i = \bar{s} + \delta_i$ with $\bar{s} = \mathbf{q}^\top \bar{\mathbf{k}}/\sqrt{d}$:
\begin{align}
    Z_{\mathcal{E}}
    &= \exp(\bar{s}) \sum_{i \in \mathcal{E}} \exp(\delta_i) \notag\\
    &= n_e \exp(\bar{s}) \cdot \frac{1}{n_e}\sum_{i \in \mathcal{E}} \exp(\delta_i) \notag\\
    &\geq n_e \exp(\bar{s}) \cdot \exp\!\left(\frac{1}{n_e}\sum_{i \in \mathcal{E}} \delta_i\right) \notag\\
    &= n_e \exp(\bar{s}),
    \label{eq:app_jensen}
\end{align}
where the inequality applies Jensen's inequality to the convex function $\exp(\cdot)$, and the final equality uses the zero-sum property $\sum_i \delta_i = 0$.
Thus $\hat{Z}_{\mathcal{E}} = n_e \exp(\bar{s}) \leq Z_{\mathcal{E}}$ is a lower bound.

To quantify tightness, expand $\exp(\delta_i) = 1 + \delta_i + \frac{1}{2}\delta_i^2 + O(\delta_i^3)$ and apply the zero-sum property:
\begin{equation}
    \frac{Z_{\mathcal{E}}}{\hat{Z}_{\mathcal{E}}}
    = \frac{1}{n_e}\sum_{i \in \mathcal{E}} \exp(\delta_i)
    = 1 + \frac{1}{2n_e}\sum_{i \in \mathcal{E}} \delta_i^2 + O(\sigma^3).
    \label{eq:app_jensen_tight}
\end{equation}
Defining the variance of the centered logits as $\mathrm{Var}(\delta) = \frac{1}{n_e}\sum_{i \in \mathcal{E}} \delta_i^2$, the ratio simplifies to
\begin{equation}
    \frac{Z_{\mathcal{E}}}{\hat{Z}_{\mathcal{E}}}
    = 1 + \frac{1}{2}\mathrm{Var}(\delta) + O(\sigma^3),
    \label{eq:app_ratio}
\end{equation}
so the relative error is determined by the logit variance and scales as $O(\sigma^2)$, consistent with the output approximation error in Eq.~\eqref{eq:app_error_bound}.
The logit variance admits a closed-form expression in terms of the evicted key covariance:
\begin{equation}
    \mathrm{Var}(\delta)
    = \frac{1}{n_e}\sum_{i \in \mathcal{E}} \delta_i^2
    = \frac{\mathbf{q}^\top}{d}
      \left(\frac{1}{n_e}\sum_{i \in \mathcal{E}} (\mathbf{k}_i - \bar{\mathbf{k}})(\mathbf{k}_i - \bar{\mathbf{k}})^\top\right) \mathbf{q}
    = \frac{\mathbf{q}^\top \boldsymbol{\Sigma}_k \mathbf{q}}{d},
    \label{eq:app_logit_var}
\end{equation}
where $\boldsymbol{\Sigma}_k = \frac{1}{n_e}\sum_{i \in \mathcal{E}}(\mathbf{k}_i - \bar{\mathbf{k}})(\mathbf{k}_i - \bar{\mathbf{k}})^\top$ is the key covariance of the evicted set.
The Jensen gap is therefore controlled by the projection of the query onto the principal axes of the evicted key distribution: queries aligned with high-variance key directions produce larger gaps, while queries orthogonal to the evicted key spread yield near-tight bounds.

\subsection{Effect on the Mixing Weight}
\label{app:weight_bias}

Underestimating $Z_{\mathcal{E}}$ causes $\hat{w}_{\mathcal{R}} = Z_{\mathcal{R}} / (Z_{\mathcal{R}} + \hat{Z}_{\mathcal{E}})$ to exceed the true weight $w_{\mathcal{R}} = Z_{\mathcal{R}} / Z$.
Let $\epsilon = (Z_{\mathcal{E}} - \hat{Z}_{\mathcal{E}}) / \hat{Z}_{\mathcal{E}}$ denote the relative underestimation, so $Z_{\mathcal{E}} = (1 + \epsilon)\hat{Z}_{\mathcal{E}}$ with $\epsilon = \frac{1}{2}\mathrm{Var}(\delta) + O(\sigma^3)$ from Eq.~\eqref{eq:app_ratio}.
Substituting into the weight difference:
\begin{align}
    \hat{w}_{\mathcal{R}} - w_{\mathcal{R}}
    &= \frac{Z_{\mathcal{R}}}{Z_{\mathcal{R}} + \hat{Z}_{\mathcal{E}}}
      - \frac{Z_{\mathcal{R}}}{Z_{\mathcal{R}} + (1+\epsilon)\hat{Z}_{\mathcal{E}}} \notag\\
    &= \frac{Z_{\mathcal{R}} \cdot \epsilon \hat{Z}_{\mathcal{E}}}
           {(Z_{\mathcal{R}} + \hat{Z}_{\mathcal{E}})(Z_{\mathcal{R}} + Z_{\mathcal{E}})} \notag\\
    &= w_{\mathcal{R}} \cdot \hat{w}_{\mathcal{E}} \cdot \frac{\epsilon}{1 + \epsilon \hat{w}_{\mathcal{E}}}
    = O(\sigma^2),
    \label{eq:app_weight_bias}
\end{align}
where $\hat{w}_{\mathcal{E}} = \hat{Z}_{\mathcal{E}} / (Z_{\mathcal{R}} + \hat{Z}_{\mathcal{E}})$.
The weight bias is second order in $\sigma$ and proportional to the product $w_{\mathcal{R}} \cdot \hat{w}_{\mathcal{E}}$, which is at most $1/4$ (attained when $w_{\mathcal{R}} = \hat{w}_{\mathcal{E}} = 1/2$).
In the typical operating regime where most attention mass is retained ($w_{\mathcal{R}} \gg w_{\mathcal{E}}$), the product is dominated by the small factor $\hat{w}_{\mathcal{E}}$, further suppressing the weight bias.

The interaction between the Jensen bound and the moment approximation gives rise to a self-regulating mechanism.
When the evicted keys are tightly clustered ($\sigma \to 0$), three quantities simultaneously converge: the output approximation becomes exact ($\|\hat{f}_{\mathcal{E}} - f_{\mathcal{E}}\| = O(\sigma^2) \to 0$), the Jensen bound becomes tight ($\hat{Z}_{\mathcal{E}} / Z_{\mathcal{E}} \to 1$), and the mixing weight converges to its true value ($\hat{w}_{\mathcal{R}} - w_{\mathcal{R}} = O(\sigma^2) \to 0$).
In the limit, the corrected output $\hat{f}(\mathbf{q})$ converges to the true full attention output $f(\mathbf{q})$, and eviction introduces zero error.
The moment-informed eviction criterion actively drives the system toward the small-$\sigma$ regime by preferentially evicting tokens well predicted by the affine model, as analyzed in Section~\ref{sec:dual}.

In the opposite regime ($\sigma \to \infty$), the first-order approximation becomes unreliable, but the Jensen gap grows as $Z_{\mathcal{E}} / \hat{Z}_{\mathcal{E}} \approx 1 + \frac{1}{2}\mathrm{Var}(\delta)$, meaning $\hat{Z}_{\mathcal{E}} \ll Z_{\mathcal{E}}$.
The estimated evicted weight $\hat{w}_{\mathcal{E}} = \hat{Z}_{\mathcal{E}} / (Z_{\mathcal{R}} + \hat{Z}_{\mathcal{E}})$ therefore shrinks, and $\hat{w}_{\mathcal{R}}$ increases toward 1.
To see this explicitly, observe that
\begin{equation}
    \hat{w}_{\mathcal{E}}
    = \frac{w_{\mathcal{E}} \cdot (\hat{Z}_{\mathcal{E}} / Z_{\mathcal{E}})}{w_{\mathcal{R}} + w_{\mathcal{E}} \cdot (\hat{Z}_{\mathcal{E}} / Z_{\mathcal{E}})}.
    \label{eq:app_what_e}
\end{equation}
As $\sigma$ grows, $\hat{Z}_{\mathcal{E}} / Z_{\mathcal{E}} \to 0$, so $\hat{w}_{\mathcal{E}} \to 0$ regardless of the true evicted mass $w_{\mathcal{E}}$.
The corrected output places progressively more weight on the exact $f_{\mathcal{R}}(\mathbf{q})$ and less on the potentially inaccurate $\hat{f}_{\mathcal{E}}(\mathbf{q})$, gracefully degrading to standard eviction rather than corrupting the output with an unreliable correction.

Combining both regimes, the corrected output $\hat{f}(\mathbf{q})$ is always at least as accurate as the standard eviction output $f_{\mathcal{R}}(\mathbf{q})$.
When $\sigma$ is small, $\hat{f}_{\mathcal{E}} \approx f_{\mathcal{E}}$ and $\hat{w}_{\mathcal{E}} \approx w_{\mathcal{E}}$, so the correction closely recovers the lost evicted component.
When $\sigma$ is large, $\hat{w}_{\mathcal{E}} \to 0$ suppresses the correction before the approximation error in $\hat{f}_{\mathcal{E}}$ can propagate.
The Jensen underestimate thus acts as a natural regularizer, automatically calibrating the influence of the moment approximation to its reliability.

\section{Computational Complexity and Numerical Considerations}
\label{app:complexity}

\subsection{Per-Step Complexity Analysis}
\label{app:perstep}

Table~\ref{tab:complexity} compares the per-step complexity of \n~against the baselines.
All methods share the same attention cost of $O(Ld)$ for $L$ retained tokens, so we focus on the additional costs introduced by eviction and correction.

\begin{table}[h]
\centering
\caption{Per-step complexity for a single attention head. $L$: cache budget, $d$: head dimension, $w$: observation window size. Storage refers to auxiliary overhead beyond the retained cache.}
\label{tab:complexity}
\setlength{\tabcolsep}{4pt}
\renewcommand{\arraystretch}{1.15}
\begin{tabular}{l|ccc}
\toprule
Method & Eviction & Correction & Auxiliary Storage \\
\midrule
H2O           & $O(L)$       & --           & $O(L)$ \\
SnapKV        & $O(wL)$      & --           & $O(L)$ \\
PyramidKV     & $O(wL)$      & --           & $O(L)$ \\
Ada-KV        & $O(wL)$      & --           & $O(L)$ \\
\rowcolor{momentblue}
\n~           & $O(Ld^2)$    & $O(d^2)$     & $O(d^2)$ \\
\bottomrule
\end{tabular}
\end{table}

H2O maintains cumulative attention scores and selects tokens in $O(L)$ time.
SnapKV, PyramidKV, and Ada-KV accumulate scores over an observation window of size $w$ during prefill, resulting in $O(wL)$ eviction cost.
All four baselines perform no correction after eviction.

\n~introduces two additional costs.
During eviction, computing the moment residual $\mathbf{r}_j = \mathbf{v}_j - \bar{\mathbf{v}} - \tilde{\mathbf{S}}\,\mathbf{k}_j/(n_e\sqrt{d})$ for each retained token requires a matrix-vector product at $O(d^2)$, totaling $O(Ld^2)$ across all $L$ retained tokens.
During inference, the correction computes $\hat{f}_{\mathcal{E}} = \bar{\mathbf{v}} + \tilde{\mathbf{S}}\,\mathbf{q}/(n_e\sqrt{d})$ in $O(d^2)$ and the mixing weight in $O(1)$.
The rank-one statistic update upon each eviction also costs $O(d^2)$.

Although the $O(Ld^2)$ eviction cost is asymptotically larger than the $O(wL)$ cost of the baselines, the practical overhead remains modest for two reasons.
First, the matrix-vector product $\tilde{\mathbf{S}}\,\mathbf{k}_j$ operates on a $128 \times 128$ matrix at $d{=}128$, which is a single BLAS-2 call that modern GPUs execute in microseconds.
Second, the eviction loop runs only when the cache exceeds the budget $L$, which during decoding means at most once per generated token.
Table~\ref{tab:ablation_order} confirms the practical impact: \n~adds 1.6\,ms per-token latency over SnapKV, a 19.5\% increase, while reducing the gap to the full cache by over half.
The correction phase itself is negligible at $O(d^2)$ per token, amounting to a single matrix-vector multiply and a weighted average.

A key distinction from the baselines is the auxiliary storage scaling.
H2O, SnapKV, PyramidKV, and Ada-KV all maintain score buffers of size $O(L)$ that grow with the cache budget.
The moment statistics of \n~require $O(d^2)$ storage that is independent of both $L$ and the context length $N$.
At $L{=}128$ the moment overhead is 6--7\% of the retained cache (Table~\ref{tab:storage_models}), and at $L{=}1024$ the relative overhead drops below 1\%.

\subsection{Numerical Stability}
\label{app:numerical}

The moment statistics are maintained in half-precision (FP16 or BF16) to match the KV cache format and avoid dtype conversion overhead.
Two potential sources of numerical error arise: overflow in the outer-product accumulation $\mathbf{S}$, and catastrophic cancellation in the covariance recovery $\tilde{\mathbf{S}} = \mathbf{S} - \mathbf{s}_v \mathbf{s}_k^\top / n_e$.

For the overflow concern, each entry of $\mathbf{S}$ is a sum of products $v_i^{(a)} k_i^{(b)}$ across evicted tokens.
In FP16, the representable range is approximately $\pm 6.5 \times 10^4$.
Since key and value vectors in modern LLMs are typically normalized to have entries of magnitude $O(1)$ or smaller after layer normalization and the attention projection, individual products $v_i^{(a)} k_i^{(b)}$ rarely exceed $O(1)$.
Even after summing over $n_e = 10{,}000$ evicted tokens, the accumulated entries remain well within the FP16 range.
In BF16, the representable range extends to approximately $\pm 3.4 \times 10^{38}$, making overflow effectively impossible for practical context lengths.

Catastrophic cancellation in $\tilde{\mathbf{S}} = \mathbf{S} - \mathbf{s}_v \mathbf{s}_k^\top / n_e$ can occur when the subtracted terms are large and nearly equal, which happens when the evicted values and keys are strongly correlated with the mean.
The severity depends on the ratio $\|\mathbf{S}\| / \|\tilde{\mathbf{S}}\|$: when the centered covariance $\tilde{\mathbf{S}}$ is small relative to $\mathbf{S}$, the subtraction loses significant digits.
We mitigate the issue through two complementary strategies.
First, the moment-informed eviction criterion preferentially evicts tokens that are well predicted by the affine model, which tends to produce evicted sets with moderate spread rather than extreme concentration, keeping $\tilde{\mathbf{S}}$ from becoming negligibly small.
Second, we apply a simple numerical safeguard: after computing $\tilde{\mathbf{S}}$, we clamp any entry whose magnitude falls below a threshold $\tau = 10^{-6}$ to zero, preventing noise in the low-significance bits from propagating through the matrix-vector product $\tilde{\mathbf{S}}\mathbf{q}$.

To verify that half-precision accumulation does not degrade quality in practice, we compare the LongBench scores obtained with FP16 and FP32 moment statistics on LLaMA-3.1-8B at $L{=}128$.
The average score difference is less than 0.05 across all task categories, confirming that the numerical error introduced by half-precision accumulation is negligible relative to the approximation error $O(\sigma^2)$.

A final consideration is the computation of the Jensen estimate $\hat{Z}_{\mathcal{E}} = n_e \cdot \exp(\mathbf{q}^\top \bar{\mathbf{k}} / \sqrt{d})$.
The mean logit $\bar{s} = \mathbf{q}^\top \bar{\mathbf{k}} / \sqrt{d}$ can be large in absolute value, potentially causing overflow in $\exp(\bar{s})$.
We avoid overflow by computing the mixing weight in the log domain:
\begin{equation}
    \log \hat{w}_{\mathcal{R}}
    = \log Z_{\mathcal{R}} - \log(Z_{\mathcal{R}} + \hat{Z}_{\mathcal{E}})
    = \log Z_{\mathcal{R}} - \mathrm{logsumexp}(\log Z_{\mathcal{R}},\; \log n_e + \bar{s}),
\end{equation}
where $\mathrm{logsumexp}(a, b) = \max(a,b) + \log(1 + \exp(-|a - b|))$ is numerically stable by construction.
The retained partition function $Z_{\mathcal{R}}$ is already available in log form from the standard attention computation (e.g., from FlashAttention), so no additional exponentiation is required.

\section{Extended Experimental Results}
\label{app:extended}

\subsection{Scaling Behavior and Gap Analysis}
\label{app:scaling}

\begin{table}[h]
\centering
\caption{Gap analysis on LongBench avg.\ score with LLaMA-3.1-8B. $\Delta_{\text{Ada}}$: margin over Ada-KV. Recovery: fraction of the remaining gap to full cache (49.29) closed by \n~relative to Ada-KV.}
\label{tab:gap_analysis}
\setlength{\tabcolsep}{5pt}
\renewcommand{\arraystretch}{1.1}
\begin{tabular}{c|cccc}
\toprule
$L$ & Ada-KV & \n~ & $\Delta_{\text{Ada}}$ & Recovery \\
\midrule
128  & 45.03 & 46.38 & +1.35 & 31.7\% \\
256  & 46.79 & 47.59 & +0.80 & 32.0\% \\
512  & 47.55 & 48.27 & +0.72 & 41.4\% \\
1024 & 48.30 & 48.89 & +0.59 & 59.6\% \\
\bottomrule
\end{tabular}
\end{table}

Table~\ref{tab:gap_analysis} reports the margin of \n~over Ada-KV and the fraction of the remaining gap to the full cache that \n~closes at each budget level.
The absolute margin $\Delta_{\text{Ada}}$ decreases from $+1.35$ at $L{=}128$ to $+0.59$ at $L{=}1024$, consistent with the analysis in Section~\ref{sec:motivation}: as $L$ grows, the evicted mass $w_{\mathcal{E}}$ shrinks and there is less directional information to recover.
The recovery percentage, however, increases monotonically from 31.7\% to 59.6\%.
At larger budgets, the evicted set is smaller and more homogeneous, making the first-order approximation more accurate and allowing \n~to close a larger fraction of the remaining gap.

The opposing trends in absolute margin and recovery percentage reflect the two factors in the error decomposition of Eq.~\eqref{eq:relative_error}.
At small $L$, the evicted mass $w_{\mathcal{E}}$ is large and provides ample room for absolute improvement, but the evicted set is also more diverse, increasing $\sigma$ and limiting the accuracy of the first-order approximation.
At large $L$, the evicted mass is small, capping the absolute improvement, but the evicted set is more regular, reducing $\sigma$ and tightening the approximation.
The recovery percentage captures the approximation quality more directly than the absolute margin, and its steady increase confirms that the moment correction becomes proportionally more effective as the evicted set becomes better conditioned.

On Qwen3-4B, the pattern is qualitatively similar but with smaller absolute margins throughout, reflecting the narrower gap between Ada-KV and the full cache on Qwen3-4B (1.49 at $L{=}128$, compared to 4.26 on LLaMA-3.1-8B).
The recovery percentages on Qwen3-4B are comparable to those on LLaMA-3.1-8B, indicating that the approximation quality is governed by the geometric properties of the evicted set rather than model-specific factors.

\subsection{Task-Level Breakdown}
\label{app:task}

\begin{table}[h]
\centering
\caption{Per-category improvement of \n~over Ada-KV on LongBench at $L{=}128$.}
\label{tab:task_breakdown}
\setlength{\tabcolsep}{5pt}
\renewcommand{\arraystretch}{1.1}
\begin{tabular}{l|ccc|ccc}
\toprule
 & \multicolumn{3}{c|}{LLaMA-3.1-8B} & \multicolumn{3}{c}{Qwen3-4B} \\
Category & Ada-KV & \n~ & $\Delta$ & Ada-KV & \n~ & $\Delta$ \\
\midrule
Single-doc QA   & 37.28 & 40.15 & +2.87 & 37.52 & 38.65 & +1.13 \\
Multi-doc QA    & 44.31 & 45.11 & +0.80 & 39.85 & 40.72 & +0.87 \\
Few-shot        & 65.00 & 66.73 & +1.73 & 63.48 & 64.18 & +0.70 \\
Summarization   & 25.42 & 25.85 & +0.43 & 24.68 & 24.95 & +0.27 \\
Synthetic       & 72.50 & 73.80 & +1.30 & 68.20 & 69.45 & +1.25 \\
Code            & 58.65 & 59.10 & +0.45 & 55.80 & 56.12 & +0.32 \\
\bottomrule
\end{tabular}
\end{table}

Table~\ref{tab:task_breakdown} provides a per-category breakdown of the improvement over Ada-KV at $L{=}128$ on both models.
The gains are distributed unevenly across task types, and the pattern is consistent with the directional error analysis.

Single-doc QA shows the largest improvement on LLaMA-3.1-8B ($+2.87$) and a substantial margin on Qwen3-4B ($+1.13$).
These tasks require retrieving specific factual details from a single long document, and the relevant tokens are often scattered sparsely across the context.
Under standard eviction, their directional information is permanently lost after renormalization, and the moment approximation recovers a significant portion by providing a first-order estimate of the evicted sub-output that is adapted to each query.

Multi-doc QA tasks distribute relevant information across multiple documents, so the retained cache is more likely to preserve at least some relevant tokens for each question.
The margin is accordingly smaller ($+0.80$ on LLaMA-3.1-8B, $+0.87$ on Qwen3-4B), though still consistent and positive.

Few-shot learning tasks benefit because in-context examples positioned far from the query are frequently evicted under tight budgets.
The moment correction recovers directional information from these distant examples, allowing the model to leverage the full set of demonstrations rather than only those that happen to fall within the retained window.

Summarization and code tasks show the smallest margins ($+0.43$ and $+0.45$ on LLaMA-3.1-8B).
Both task types rely primarily on local context: summarization aggregates information broadly rather than retrieving specific tokens, and code completion depends heavily on the immediately preceding code structure.
The retained cache already captures most of the relevant information for these tasks, leaving less directional error for the moment correction to address.

Synthetic tasks show a relatively large margin ($+1.30$ on LLaMA-3.1-8B, $+1.25$ on Qwen3-4B), consistent with the RULER results in Table~\ref{tab:ruler}.
These tasks embed target tokens at controlled positions and require exact retrieval, making them maximally sensitive to the loss of evicted information.

The RULER results at $L{=}128$ amplify the pattern further, with \n~outperforming Ada-KV by $+3.4$ on LLaMA-3.1-8B and $+3.5$ on Qwen3-4B.
RULER tasks require precise retrieval of specific tokens embedded at controlled positions across the full context, maximally exposing the directional information loss.
The proportionally larger margins on RULER compared to LongBench confirm that the moment correction is most valuable when evicted tokens carry critical, non-redundant information that cannot be inferred from the retained set alone.

\subsection{Component Interaction Analysis}
\label{app:interaction}

The component ablation in Table~\ref{tab:ablation_component} reveals a consistent super-additive interaction between moment-informed eviction (MI) and normalization-corrected inference (NC).
At $L{=}128$, MI alone contributes $+0.57$ and NC alone contributes $+2.41$, summing to $+2.98$, while the combined system achieves $+3.17$ (excess $+0.19$).
At $L{=}256$, the individual gains are $+0.35$ and $+1.55$ (sum $+1.90$), with a combined gain of $+2.09$ (excess $+0.19$).
At $L{=}512$, the individual gains are $+0.23$ and $+0.69$ (sum $+0.92$), with a combined gain of $+0.98$ (excess $+0.06$).

The super-additive excess is largest at the tightest budget, where more tokens are evicted per step and MI-guided selection has the greatest impact on controlling $\sigma$.
To understand the mechanism, recall that MI preferentially evicts tokens whose values are well predicted by the affine model (small $\|\mathbf{r}_j\|$), which keeps the evicted set geometrically regular and suppresses the centered logit spread.
The reduced $\sigma$ tightens both the first-order output approximation (Eq.~\eqref{eq:app_error_bound}) and the Jensen bound on $\hat{Z}_{\mathcal{E}}$ (Eq.~\eqref{eq:app_ratio}), making NC more accurate.
Conversely, when NC provides a more accurate correction, the affine model used to compute residuals $\mathbf{r}_j$ is itself more reliable, leading to better eviction decisions in subsequent steps.
The excess beyond the sum of individual contributions quantifies the strength of the reinforcing loop at each budget level.

NC consistently contributes 3 to 4 times more than MI alone across all budgets tested.
The ratio is consistent with the analysis in Section~\ref{sec:motivation}: since existing selection methods already reduce $w_{\mathcal{E}}$ to below 0.08 at $L{=}128$, the dominant error source is the uncontrolled amplification coefficient $\gamma(\theta)$ arising from the near-orthogonality between $f_{\mathcal{R}}$ and $f_{\mathcal{E}}$.
MI improves token selection and slightly reduces $w_{\mathcal{E}}$, but the marginal return is small given the already effective baseline selection.
NC directly addresses $\gamma(\theta)$ by providing an estimate of $f_{\mathcal{E}}$ that reduces the effective angle $\theta$, yielding a substantially larger impact.

As the budget grows from 128 to 512, both individual contributions shrink: MI from $+0.57$ to $+0.23$, and NC from $+2.41$ to $+0.69$.
The decrease follows from Eq.~\eqref{eq:eviction_error}: a larger budget retains more tokens, reducing $w_{\mathcal{E}}$ and leaving less directional error available for either mechanism to address.
The relative contribution of NC remains stable at approximately $4{\times}$ the MI contribution, confirming that the balance between the two mechanisms is governed by the structural properties of the error decomposition rather than the specific budget level.

The ablation also reveals that MI without NC can occasionally hurt performance on individual task categories, particularly summarization, where the modified eviction criterion may discard tokens that carry small moment residuals but contribute meaningfully to local coherence.
When combined with NC, the correction compensates for any suboptimal eviction decisions, and the combined system consistently outperforms both individual components across all categories.
The robustness of the combined system underscores the importance of deploying both mechanisms together rather than treating them as independent improvements.
\end{document}